\documentclass[runningheads]{llncs}

\usepackage{eccv}

\usepackage{enumitem}
\usepackage{multirow}
\usepackage{colortbl}
\usepackage{microtype}
\usepackage{comment}
\usepackage[hang,flushmargin]{footmisc}

\usepackage{amsfonts}
\usepackage{tcolorbox}

\setlist[itemize]{align=parleft,left=0pt,topsep=1mm,itemsep=0mm,parsep=1mm}

\definecolor{postechred}{rgb}{0.784, 0.003, 0.313}

\definecolor{ballblue}{rgb}{0.13, 0.67, 0.8}
\definecolor{cornellred}{rgb}{0.7, 0.11, 0.11}
\definecolor{darkcyan}{rgb}{0.0, 0.55, 0.55}
\definecolor{CuGray}{gray}{0.9}
\definecolor{airforceblue}{rgb}{0.36, 0.54, 0.66}
\definecolor{rev}{rgb}{0.784, 0.003, 0.313}
\definecolor{pink}{cmyk}{0, 0.7808, 0.4429, 0.1412}
\definecolor{amethyst}{rgb}{0.6, 0.4, 0.8}
\definecolor{black}{rgb}{0.0, 0.0, 0.0}
\definecolor{dimgray}{rgb}{0.41, 0.41, 0.41}
\definecolor{bleudefrance}{rgb}{0.19, 0.55, 0.91}
\definecolor{blue(ryb)}{rgb}{0.01, 0.28, 1.0}
\definecolor{Gray}{gray}{0.88}
\definecolor{green(ncs)}{rgb}{0.0, 0.62, 0.42}
\definecolor{brightpink}{rgb}{1.0, 0.0, 0.5}
\definecolor{alizarin}{rgb}{0.82, 0.1, 0.26}
\definecolor{orange-red}{rgb}{1.0, 0.27, 0.0}
\definecolor{nicegreen}{rgb}{0.0, 0.7, 0.1}
\definecolor{clova}{rgb}{0.24, 0.63, 0.33}

\definecolor{kellygreen}{rgb}{0.3, 0.73, 0.09}

\newcolumntype{g}{>{\columncolor{CuGray}}c}
\newcolumntype{z}{>{\columncolor{CuGray}}l}

\renewcommand{\paragraph}[1]{\vspace{1mm}\noindent\textbf{#1.}\,}

\usepackage{xspace}

\makeatletter
\def\@fnsymbol#1{\ensuremath{\ifcase#1\or *\or \dagger\or \ddagger\or
   \mathsection\or \mathparagraph\or \|\or **\or \dagger\dagger
   \or \ddagger\ddagger \else\@ctrerr\fi}}
\makeatother

\def\onedot{.\@\xspace}

\newcommand{\Sref}[1]{Sec.~\ref{#1}}

\newcommand{\Fref}[1]{Fig.~\ref{#1}}
\newcommand{\Tref}[1]{Table~\ref{#1}}

\newcommand{\Real}{\mathbb R}

\newcommand{\be}{\begin{eqnarray}}
\newcommand{\ee}{\end{eqnarray}}
\newcommand{\bee}{\begin{eqnarray*}}
\newcommand{\eee}{\end{eqnarray*}}

\newcommand{\matrixb}{\left[ \begin{array}}
\newcommand{\matrixe}{\end{array} \right]}

\usepackage{amssymb}
\usepackage{pifont}
\usepackage{algorithm}
\usepackage{algpseudocode}

\usepackage{makecell}
\newcommand{\nickname}{SplatReasoner\xspace}

\definecolor{gu}{rgb}{0.5460, 0.1755, 0.2766}
\definecolor{clova}{rgb}{0.24, 0.63, 0.33}
\definecolor{dahye}{rgb}{1.0, 0.55, 0.0}
\definecolor{skyblue}{rgb}{0.20, 0.55, 0.80}
\definecolor{softred}{rgb}{0.90, 0.30, 0.30}

\makeatletter
\DeclareRobustCommand\onedot{\futurelet\@let@token\@onedot}
\def\@onedot{\ifx\@let@token.\else.\null\fi\xspace}

\newcommand{\best}[1]{\textbf{#1}}

\usepackage{eccvabbrv}

\usepackage{graphicx}
\usepackage{booktabs}

\usepackage[accsupp]{axessibility}  

\usepackage{multirow}
\usepackage{subcaption}
\usepackage{wrapfig}
\usepackage[table,xcdraw]{xcolor}

\definecolor{oursblue}{HTML}{D2E3FC}
\definecolor{oursgreen}{HTML}{C8E6C9}
\definecolor{oursgray}{HTML}{E8E8E8}

\usepackage{hyperref}

\usepackage{orcidlink}

\begin{document}

\title{\nickname: Enhancing Embodied Reasoning and Grounding by Novel View Synthesis}


\titlerunning{\nickname}

\author{
Kim Yu-Ji\inst{1}\orcidlink{0000-0002-1774-5442} \quad
Dahye Lee\inst{2}\orcidlink{0009-0003-8125-284X} \quad
Kim Jun-Seong\inst{1}\orcidlink{0000-0001-7570-6508} \quad
Nam Hyeon-Woo\inst{1}\orcidlink{0000-0001-9543-3770} \quad
GeonU Kim\inst{2}\orcidlink{0009-0009-0224-0060} \quad
Yongjin Kwon\inst{3}\orcidlink{0000-0001-8818-4657} \quad
Yu-Chiang Frank Wang\inst{4}\orcidlink{0000-0002-2333-157X} \\
Jaesung Choe\inst{4}\orcidlink{0000-0002-8978-6702} \quad
Tae-Hyun Oh\inst{2}\orcidlink{0000-0003-0468-1571}
}

\authorrunning{K.~Yu-Ji et al.}

\institute{
{$^{1}$POSTECH} \quad
{$^{2}$KAIST}  \quad
{$^{3}$ETRI}  \quad
{$^{4}$NVIDIA}  \\
\email{ugkim@postech.ac.kr} \\
\url{https://splatreasoner.github.io/}
}

\maketitle

\begin{abstract}

Vision-Language Models (VLMs) have demonstrated strong reasoning capabilities over images and videos, yet their application to embodied scene understanding often constrained by the fixed viewpoints stored in episodic RGB-D memories.
These observations may fail to capture query-relevant evidence due to occlusions, object truncation, restricted fields of view, or suboptimal view composition.
We present \nickname, a framework that introduces novel view synthesis into the VLM reasoning process by leveraging 3D Gaussian Splatting (3DGS).
Given a user query about a 3D scene, \nickname retrieves relevant observations and synthesizes query-conditioned viewpoints that reveal the visual evidence needed to answer the query and ground the referred entities in 3D.
Experiments show that query-conditioned novel view synthesis improves both embodied reasoning and 3D grounding over fixed-view memory and language-embedded 3DGS baselines.

\keywords{Embodied Reasoning \and 3D Grounding \and Novel View Synthesis}

\end{abstract}

\section{Introduction}
\label{sec:intro}

\begin{figure}[t]
    \centering
    \includegraphics[width=1.0\linewidth]{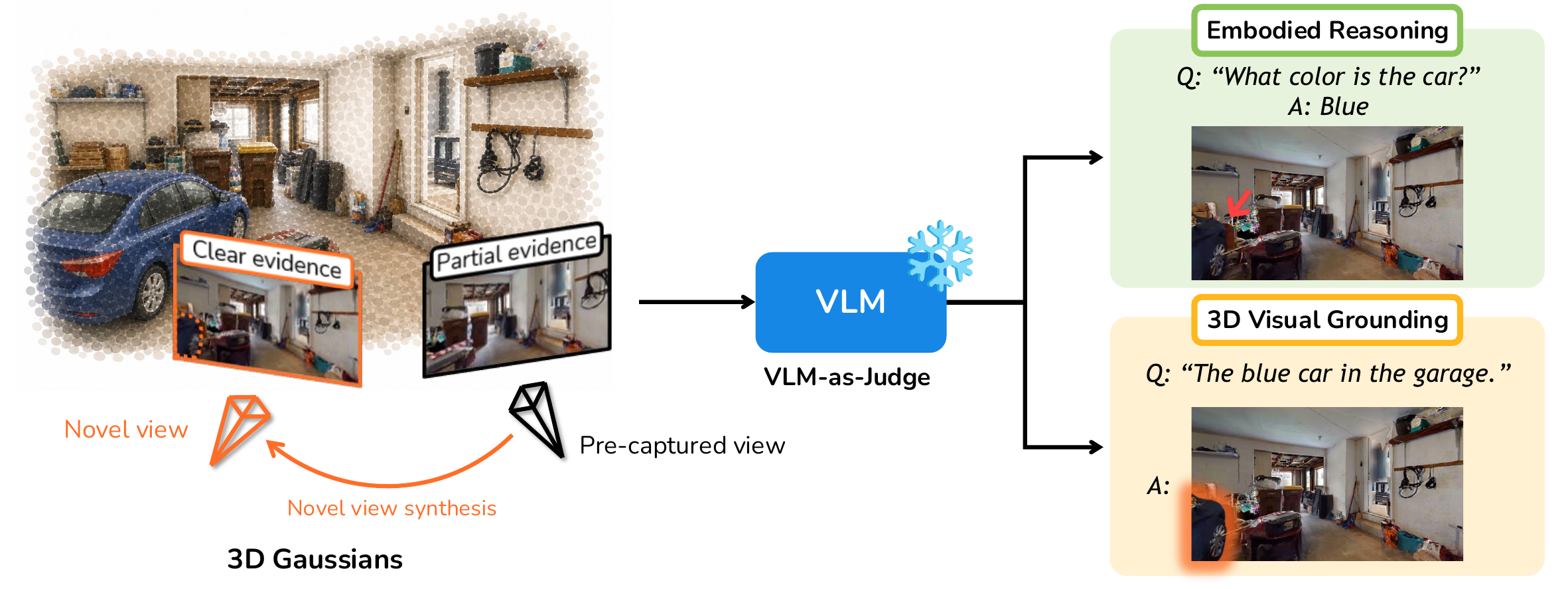}
    \caption{
        \textbf{\nickname} aims to enhance embodied task capacity by integrating VLM with novel view synthesis of the 3DGS representation.
        Specifically, \nickname retrieves and synthesizes informative camera viewpoints to provide the most relevant visual evidence, thereby facilitating improved query answering and 3D entity grounding.
    }
    \vspace{-4mm}
    \label{fig:teaser}
\end{figure}

Building embodied agents capable of understanding 3D environments~\cite{lei2025gaussnav} and executing natural-language instructions~\cite{chen2020scanrefer, zhang2023multi3drefer} remains a foundational goal in computer vision and robotics. Recently, Vision-Language Models (VLMs)~\cite{hurst2024gpt4o, liu2024llavaimproved,bai2025qwen3,regiongpt,spatialrgpt} have demonstrated advanced reasoning and grounding capability for semantic understanding. As these models have proven effective at interpreting image and video data, research has naturally pivoted toward applying their capabilities to embodied tasks in 3D scenes. However, bridging this gap is a critical challenge, as it requires equipping 2D-native VLMs with a cohesive memory of past observations and a robust, structural understanding of 3D space.

Previous approaches~\cite{3d-mem,gu2024conceptgraphs, das2018embodied, majumdar2024openeqa, cangea2019videonavqa, regiongpt,spatialrgpt} demonstrate that VLMs can perform complex reasoning and grounding across multi-view images when provided with structured memory built from RGB-D observations. While these approaches make notable progress toward memory-based embodied reasoning, they typically operate on a set of pre-captured image sequences as a structured memory and object-level abstractions, which naturally constrain the investigation of 3D environments into a fixed set of 2D views.
Relying on a fixed set of pre-recorded images inherently restricts a VLM's reasoning and grounding capabilities. Consequently, these models often struggle with sub-optimal view compositions, such as occlusions, small objects, and truncations at image boundaries.
While the state-of-the-art methods, including recent 3D-Mem~\cite{3d-mem}, demonstrated strong reasoning capabilities, they particularly faces challenges in these cases by design.

In this context, 3D Gaussian Splatting (3DGS)~\cite{3dgs} can be an advantageous representation alternative to the fixed-view based memory by virtue of its favorable novel view synthesis functionality.
As a separate line of research, language-driven 3D scene understanding methods~\cite{langsplat, legaussians, wu2024opengaussian, refersplat, drsplat, marrie2025ludvig, an2025c3g, lee2026embodiedsplat, ding2026extrinsplat, xiong2025splat} have been proposed to connect 3D and language with the novel view synthesis function, enabling open-vocabulary 3D scene understanding.
These approaches also successfully demonstrate the effectiveness in alignment between language and 3D representations.
However, their similarity computation strategies are primarily optimized for simple, category-level text queries (e.g., ``chair'' or ``table'').
Consequently, it remains challenging to extend them to support the complex reasoning and grounding tasks required for embodied applications.

To address these bottlenecks, we introduce \nickname, a framework that performs VLM-guided embodied reasoning and grounding based on novel view synthesis of 3DGS, as shown in~\Fref{fig:teaser}.
For a given query, \nickname (1) selects initial candidate views by matching simplified queries to 3D Gaussians, and (2) conducts a VLM-as-Judge mechanism with synthesized novel views to finalize the view selection for downstream tasks.
Instead of relying on pre-defined fixed views, our pipeline introduces 3D Gaussian novel view synthesis to improve VLM-guided reasoning and grounding for embodied tasks.
The contributions of our paper are summarized as follows:
\begin{itemize}
    \item We introduce \nickname, a novel framework enhancing VLM-guided embodied reasoning and grounding by integrating novel view synthesis of 3DGS.
    \item We demonstrate the critical importance of VLM-as-Judge with novel view synthesis for spatial reasoning tasks, effectively overcoming the strict viewpoint limitations of existing fixed-view memory approaches.
    \item We achieve strong performance improvements on embodied question answering, and we demonstrate that our enhanced reasoning pipeline enables precise 3D visual grounding.
\end{itemize}
 
\section{Related Work}
\label{sec:related_work}

\paragraph{Embodied Reasoning with 2D VLMs}
To perform embodied reasoning tasks~\cite{das2018embodied, majumdar2024openeqa, cangea2019videonavqa}, recent approaches~\cite{llava, liu2024llavaimproved, openai2023gpt4v, hurst2024gpt4o, singh2025openaigpt5, bai2025qwen3} integrate VLMs with a structured memory built from RGB or RGB-D~\cite{gu2024conceptgraphs, majumdar2024openeqa, 3d-mem} observations.
Specifically, a text-centric method generates captions\cite{changpinyo-etal-2022-may} from RGB data, which are then fed to LLMs~\cite{touvron2023llama, openai2023gpt4} that reason without direct visual context. 
In contrast, structured abstraction methods build scene graphs from RGB-D data~\cite{gu2024conceptgraphs, zhang2025open, saxena2025grapheqa, kimtopological, yin2024sg}, representing objects as nodes and relationships as edges. 
Nevertheless, a significant limitation of both approaches is their reliance on high-level abstraction. 
By discarding low-level geometric and visual details, these methods hinder the fine-grained understanding required for complex tasks.

While the development of VLMs has enabled the use of visually rich, multi-view images, this introduces a new challenge: long visual sequences often overwhelm models and limit effective reasoning~\cite{wang2025mmlongbenchbenchmarkinglongcontextvisionlanguage, ge2025v2pe, sharma2024losing}. 
3D-Mem~\cite{3d-mem} offers a solution by compressing these observations into ``Memory Snapshots.'' This method removes visual redundancy and stores object-level annotations. 
Consequently, VLMs can access only the most relevant snapshots for a given query, allowing them to reason efficiently while maintaining the visual information needed for complex question answering and spatial reasoning. A primary bottleneck of these memory-based systems is their reliance on pre-captured, static views. While they may exhibit strong semantic reasoning, they fall short on tasks that demand 3D localization, novel view exploration, or precise spatial grounding.

\paragraph{Language-embedded 3D Gaussian  (3DGS)}
3DGS~\cite{3dgs} is a modern 3D representation that enables fast, high-fidelity 3D reconstruction and real-time rendering. 
It has evolved beyond rendering, as several works~\cite{langsplat, drsplat, legaussians, wu2024opengaussian, refersplat, zhou2024feature, ji2025fastlgs, ye2024gaussian, marrie2025ludvig, an2025c3g, lee2026embodiedsplat, ding2026extrinsplat, xiong2025splat} have augmented it with language features for open-vocabulary reasoning. 
For example, prior approaches~\cite{langsplat, drsplat, legaussians, wu2024opengaussian} have embedded CLIP features~\cite{clip} into each Gaussian, enabling open-vocabulary object retrieval and segmentation.

The core challenge with existing embedding-augmented 3DGS approaches is that they primarily rely on static embedding similarity. 
While effective for simple category-level queries, they struggle to capture richer linguistic reasoning, such as compositional instructions.
For example, given the query ``What color are pillows in the kitchen?'', such methods may retrieve all pillows in the scene, rather than those specifically associated with the kitchen.
This behavior arises because their visual reasoning operates at the level of individual Gaussians, without explicit mechanisms for iterative grounding or contextual aggregation.

More recently, ReferSplat~\cite{refersplat} aligns fine-grained textual descriptions with Gaussian features, leading to a deeper understanding of spatial relationships.
However, because its feature updates are rendering-based, it remains less suitable for direct 3D search~\cite{drsplat}.
By coupling VLM reasoning~\cite{llava, liu2024llavaimproved, openai2023gpt4v, hurst2024gpt4o, singh2025openaigpt5, bai2025qwen3} with a modern 3D representation, specifically 3DGS~\cite{3dgs}, our method enables direct search and grounding in 3D space.
Our approach also allows agents to actively synthesize novel viewpoints, disambiguate queries, and ground targets with high spatial accuracy through advanced VLM reasoning.

\paragraph{Embodied Reasoning with 3D VLMs}
Recent advances in embodied reasoning have introduced 3D VLMs~\cite{hong20233d, huang2024embodied, xu2024pointllm, huang2024chat} that process 3D representations directly~\cite{chen2020scanrefer, azuma2022scanqa, ma2022sqa3d, zhang2023multi3drefer}. 
While recent studies have integrated 3DGS into VLMs~\cite{halacheva2025gaussianvlm}, these methods still require additional fine-tuning. In contrast, our approach enables 2D pre-trained models to perceive and reason in 3D environments without further training by leveraging rich 2D VLMs.

\section{Method}
\label{sec:method}

\begin{figure}[!t]
    \centering
    \includegraphics[width=1.0\linewidth]{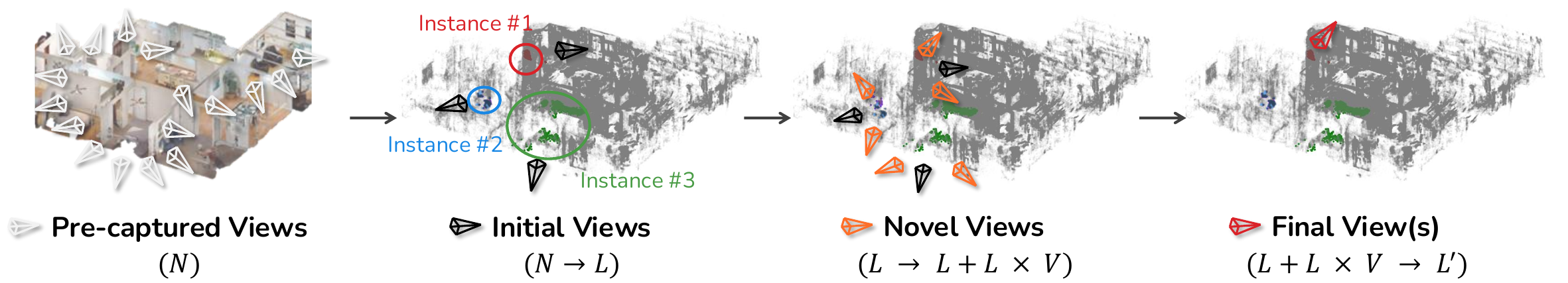}
    \caption{\textbf{Overview.} 
    Given $N$ pre-captured views, we sample $L'$ images to perform embodied tasks. 
    First, we select $L$ initial views from $N$ input views through initial view selection ($N \gg L$).
    Then, we augment each initial view with $V$ novel views.
    Finally, we proceed with the final view selection with VLMs to find most informative views to answer the query.
    }
    \vspace{-4mm}
    \label{fig:overview}
\end{figure}

Given a user query $\mathcal{Q}$ about a 3D scene, our goal is to retrieve and synthesize the camera viewpoints that provide the most relevant visual evidence for answering the query and grounding the referred entities in 3D.
The scene is observed through $N$ posed RGB images $\mathcal{I}=\{I_i\}_{i=1}^{N}$ with camera poses $\mathcal{H}=\{H_i\}_{i=1}^{N}$, where each pose $H_i=[\mathbf{R}_i|\mathbf{t}_i]\in\Real^{3\times4}$ denotes the extrinsic matrix composed of rotation $\mathbf{R}_i$ and translation $\mathbf{t}_i$.

An overview of our framework is shown in \Fref{fig:overview}.
To balance efficiency and evidence quality, we decompose this process into initial view selection (\Sref{subsec:best_view}), which retrieves visible query-relevant regions from $N$ input views to $L$ selected views ($N \gg L$). 
Then, the system performs novel view synthesis (\Sref{subsec:novel_view}) and conducts final view selection (\Sref{subsec:vlm_as_judge}), which compares the initial and synthesized novel views to select the most informative evidence and verify whether it is sufficient to answer the query.
The selected final views are utilized for downstream tasks (\Sref{subsec:downstream_task}), such as embodied question answering and 3D visual grounding.

\begin{figure}[!t]
    \centering
    \includegraphics[width=1.0\linewidth]{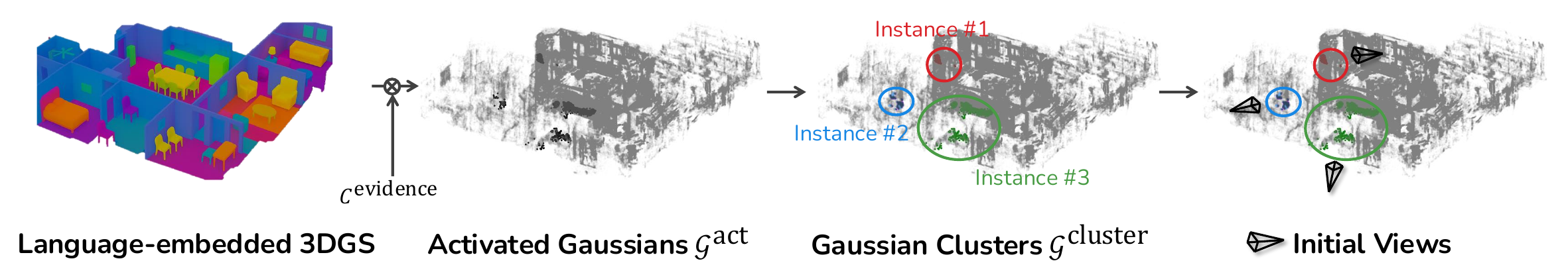}
    \caption{\textbf{Initial view selection.} 
    First, we identify the activated Gaussians $\mathcal{G}^{\mathrm{act}}$ using a set of evidence categories $\mathcal{C}^{\mathrm{evidence}}$ extracted from the query $\mathcal{Q}$. 
    Next, these activated Gaussians are grouped into clusters $\{\mathcal{G}_l^{\mathrm{cluster}}\}_{l=1}^{L}$ that serve as instance-level candidates. Finally, the initial viewpoints are determined by maximizing the visibility score.
    }
    \vspace{-4mm}
    \label{fig:initial_view_selection}
\end{figure}

\subsection{Initial View Selection in Language-embedded 3D Gaussians}
\label{subsec:best_view}

We first localize the query-relevant region in 3D and then search for viewpoints that maximize its visibility as shown in~\Fref{fig:initial_view_selection}.
By measuring visibility in 3D rather than in 2D pixel space, our approach avoids biases from occlusion and camera distance, enabling occlusion-aware viewpoint selection.

\paragraph{Construction of language-embedded 3D Gaussians}
We first reconstruct the scene as a set of $M$ 3D Gaussians,
\begin{equation}
    \mathcal{G}=\{G_j\}_{j=1}^{M},
\end{equation}
where each Gaussian $G_j$ contains its geometric and appearance parameters from 3DGS, such as position, scale, opacity, and color.
To use this representation as a queryable memory, we additionally attach semantic information to each Gaussian.
Depending on the downstream task and the comparison protocol, this semantic attribute is represented in one of two forms, as described below.

For embodied reasoning, we follow 3D-Mem~\cite{3d-mem} and use a closed-set object detector~\cite{varghese2024yolov8} to assign each Gaussian a category label $C_j\in\mathcal{C}$, where $\mathcal{C}$ denotes the detector vocabulary.
This category-level memory is suitable for retrieving common evidence objects, such as \textit{pillow} or \textit{cushion}, from natural language questions.
For 3D object localization and 3D referring segmentation, we adopt the open-set pipeline of Dr. Splat~\cite{drsplat}.
Specifically, we segment the input images with SAM~\cite{sam}, extract a CLIP~\cite{clip} feature for each segment, and register these features to the reconstructed Gaussians.
The resulting Gaussian $G_j$ is associated with an embedding $f_j$ that can be compared with text embeddings.
In both settings, we use the direct registration strategy from~\cite{drsplat} to aggregate multi-view semantic observations into the 3DGS representation, yielding a compact structured memory that can be searched directly in 3D space.

\paragraph{Gaussian search via query-driven evidence categories}
The initial view selection stage converts the user query into a small set of 3D candidate regions and chooses one pre-recorded view for each region.
This stage reduces the number of images passed to VLMs while preserving the visual evidence that is likely to answer $\mathcal{Q}$.
It consists of three steps: extracting evidence categories from the query, clustering the activated Gaussians into instance-level candidates, and selecting the most visible input view for each cluster.

Natural language questions often contain reasoning intents that are not directly usable as 3D search keys.
We therefore first use a Large Language Model (LLM) to extract a set of evidence categories $\mathcal{C}^{\mathrm{evidence}}\subseteq\mathcal{C}$ from the query.
These categories are objects or semantic concepts that are likely to provide evidence for answering the query.
For example:
\begin{tcolorbox}[colback=gray!5, colframe=gray!75]
\resizebox{0.7\linewidth}{!}{
\begin{tabular}{l}
    \textbf{Prompt:} You should retrieve helpful objects in order.  \\
    \textbf{Query:} Where can I take a nap? \\
    \textbf{Categories ($\mathcal{C}$):} radiator, cushion, sink, pillow, picture, ... \\
    \textbf{Evidence categories ($\mathcal{C}^{\mathrm{evidence}}$):} pillow, cushion.
\end{tabular}}
\label{textbox}
\end{tcolorbox}

\noindent
For embodied reasoning, each Gaussian has a closed-set category label $C_j$.
We therefore activate the Gaussians whose labels match the selected evidence categories:
\begin{equation}
    \mathcal{G}^{\mathrm{act}}
    =
    \left\{
        G_j \in \mathcal{G}
        \;\middle|\;
        C_j \in \mathcal{C}^{\mathrm{evidence}}
    \right\}.
\end{equation}
where $\mathcal{G}^{\mathrm{act}}$ denotes the set of query-activated Gaussians.


For 3D grounding, where each Gaussian is represented by an open-set feature
$f_j$, we independently activate Gaussians for each evidence category.
Let $t_c$ denote the text embedding of category
$c\in\mathcal{C}^{\mathrm{evidence}}$, and let
$\mathrm{sim}(\cdot,\cdot)$ denote cosine similarity.
We combine all category-specific activations as
\begin{equation}
    \mathcal{G}^{\mathrm{act}}
    =
    \bigcup_{c\in\mathcal{C}^{\mathrm{evidence}}}
    \left\{
        G_j\in\mathcal{G}
        \;\middle|\;
        \mathrm{sim}(f_j,t_c)\ge\tau
    \right\}.
\end{equation}
We set $\tau=0.5$ for 3D grounding.

\paragraph{Clustering 3D Gaussians for instance assignment}
A single Gaussian is a local primitive and does not by itself represent an entire object.
We therefore group activated Gaussians into spatial clusters that serve as instance-level candidates.
Let $\mathbf{x}_j\in\Real^3$ denote the 3D center of Gaussian $G_j$.
We cluster $\mathcal{G}^{\mathrm{act}}$ according to the spatial distribution of $\{\mathbf{x}_j\}$ using HDBSCAN~\cite{dbscan, hdbscan}:
\begin{equation}
    \{\mathcal{G}_l^{\mathrm{cluster}}\}_{l=1}^{L}
    =
    \operatorname{Cluster}(\mathcal{G}^{\mathrm{act}}),
\end{equation}
where $\mathcal{G}_l^{\mathrm{cluster}}$ is the $l$-th Gaussian cluster and $L$ is the number of clusters adaptively determined by the algorithm.
When category labels are available, clustering is applied within the same semantic category to avoid grouping different object types.
Since clustering is used to reduce the number of candidate viewpoints for VLM processing, coarse grouping is sufficient even if nearby parts of the same semantic region are merged.

\paragraph{Initial view selection with visibility score}
Given $L$ Gaussian clusters $\{\mathcal{G}^{\mathrm{cluster}}_l\}_{l=1}^{L}$, we select one initial input pose for each cluster from $\mathcal{H}$.
The selected initial pose set is denoted as
\begin{equation}
    \mathcal{H}^{\mathrm{init}}
    =
    \{H_l^{\mathrm{init}}\}_{l=1}^{L},
\end{equation}
where $H_l^{\mathrm{init}}$ is the input view chosen for cluster $\mathcal{G}^{\mathrm{cluster}}_l$.
For each pair of cluster $\mathcal{G}^{\mathrm{cluster}}_l$ and input pose $H_i$, we render a cluster activation map using the splatting renderer~\cite{3dgs, zwicker2001ewa}.
Let $\mathcal{M}_{l,i}^{\mathrm{visible}}$ be the set of pixels with positive activation in this rendered map.

To account for occlusion, we do not simply count projected Gaussians.
Instead, for every pixel $\mathbf{u}\in \mathcal{M}_{l,i}^{\mathrm{visible}}$, we identify the Gaussian in the full scene that contributes the maximum rendering weight:
\begin{equation}
    \mathcal{G}^{\mathrm{max}}_{l,i}
    =
    \left\{
    G_{j^\ast}
    \;\middle|\;
    \mathbf{u} \in \mathcal{M}_{l,i}^{\mathrm{visible}},
    \ j^\ast =
    \operatorname*{arg\,max}_{m \in \{1,\dots,M\}}
    w_m(H_i, \mathbf{u}; G_m)
    \right\}.
\end{equation}
Here, $w_m(H_i,\mathbf{u};G_m)$ is the rendering weight of Gaussian $G_m$ at pixel $\mathbf{u}$ under camera pose $H_i$.
If a target cluster is occluded, the dominant Gaussian at the corresponding pixel is likely to come from a foreground object rather than from the target cluster.
The visible subset of cluster $\mathcal{G}^{\mathrm{cluster}}_l$ is therefore defined as
\begin{equation}
    \mathcal{G}^{\mathrm{vis}}_{l,i}
    =
    \mathcal{G}^{\mathrm{max}}_{l,i}
    \cap
    \mathcal{G}^{\mathrm{cluster}}_l.
\end{equation}
We then define the visibility score $S^{\mathrm{vis}}$ as the fraction of cluster Gaussians that are visible from $H_i$:
\begin{equation}
    S^{\mathrm{vis}}_{l,i}
    =
    \frac{|\mathcal{G}^{\mathrm{vis}}_{l,i}|}
    {|\mathcal{G}^{\mathrm{cluster}}_l|}.
\end{equation}
Finally, the initial view for cluster $l$ is selected by
\begin{equation}
    H_l^{\mathrm{init}} = H_{i_l^\ast},
    \quad
    i_l^\ast = \operatorname*{arg\,max}_{i \in \{1,\dots,N\}} S^{\mathrm{vis}}_{l,i}.
\end{equation}
The visibility-based selection favors views where the queried instance is not merely projected into the image but is actually visible after considering occluding Gaussians.

\subsection{Novel View Synthesis}
\label{subsec:novel_view}

Unlike previous study~\cite{3d-mem}, which are constrained to pre-recorded images for visual reasoning, our model integrates novel view synthesis directly into the reasoning pipeline.
This pipeline is effective when the initial camera views are suboptimal due to scene ambiguities such as occlusions or complex object relationships. 

We therefore perturb the initial viewpoints via novel-view rendering. 
We generate perturbed initial views with novel view synthesis and employ a VLM-as-Judge~\cite{hurst2024gpt4o, bai2025qwen3} module to assess their informativeness, ultimately selecting more suitable viewpoints for the given user queries.


Given the initial views
$\mathcal{H}^{\mathrm{init}}
=\{H_l^{\mathrm{init}}\}_{l=1}^{L}$,
we synthesize $V$ novel views around each initial pose:
\begin{equation}
    \mathcal{H}^{\mathrm{novel}}_l
    =
    \{\widetilde{H}_{l,v}\}_{v=1}^{V},
    \qquad l=1,\ldots,L.
\end{equation}
For each initial pose, we define its local candidate group by combining the
initial pose with its synthesized novel poses:
\begin{equation}
    \mathcal{H}^{\mathrm{candi}}_l
    =
    \{H_l^{\mathrm{init}}\}
    \cup
    \mathcal{H}^{\mathrm{novel}}_l.
\end{equation}
The complete collections are
$\mathcal{H}^{\mathrm{novel}}
=\bigcup_{l=1}^{L}\mathcal{H}^{\mathrm{novel}}_l$
and
$\mathcal{H}^{\mathrm{candi}}
=\bigcup_{l=1}^{L}\mathcal{H}^{\mathrm{candi}}_l$.
We set $V=4$, corresponding to the left, right, forward, and backward
perturbations, resulting in $L\times V$ synthesized novel views.

\begin{figure}[!t]
    \centering
    \includegraphics[width=1.0\linewidth]{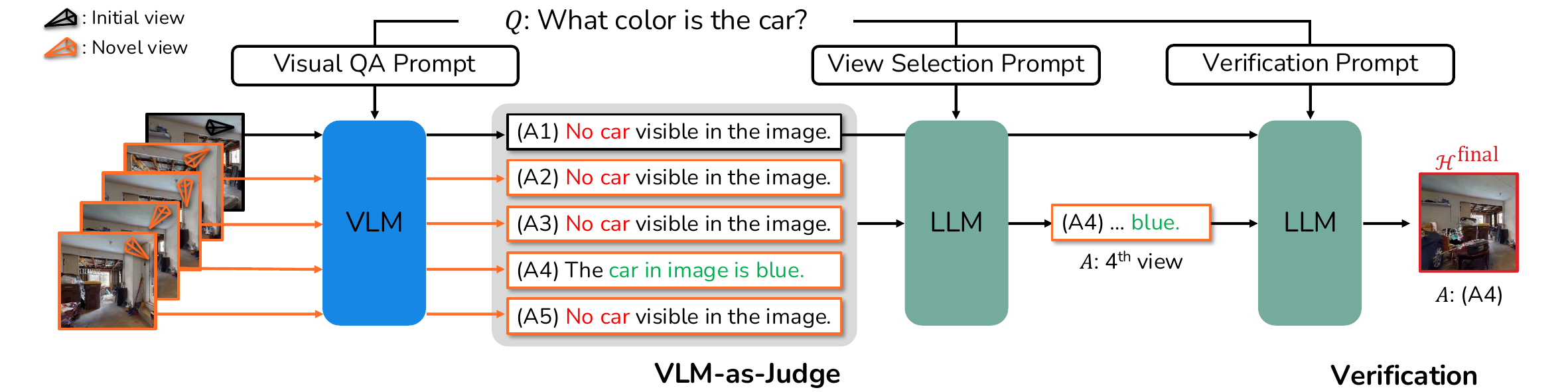}
    \caption{\textbf{Final view selection.}
    For each initial view, we synthesize novel views and employ a VLM-as-Judge to select the most informative viewpoints from both the initial and novel view candidates. 
    To ensure that the selected view from VLM-as-Judge provide richer information compared to the initial one, we conduct a verification stage.
    }
    \vspace{-4mm}
    \label{fig:final_view_selection}
\end{figure}

\subsection{Final View Selection}\label{subsec:vlm_as_judge}
We employ VLMs for the final 3D view selection because visibility scores alone are insufficient to measure the informativeness of novel views.
As illustrated in~\Fref{fig:final_view_selection}, VLMs serve as judges for both novel view adjustments and verification.

\paragraph{VLM-as-Judge for estimating view informativeness} 
%
Final-view selection is performed independently within each local candidate
group $\mathcal{H}_l^{\mathrm{candi}}$ for $l=1,\ldots,L$.
Each candidate view in the group is evaluated independently by feeding its
rendered image and the query to a VLM~\cite{hurst2024gpt4o,bai2025qwen3},
because VLMs tend to struggle with long visual
sequences~\cite{wang2025mmlongbenchbenchmarkinglongcontextvisionlanguage,
ge2025v2pe,sharma2024losing,park2025mitigating,lee2026enhancing}.

The VLM converts the visual evidence from each candidate view into a textual
answer candidate.
The LLM then compares the textual candidates within the same group and
selects the answer that most directly addresses the query, following the
text-based aggregation strategy in~\cite{zeng2022socratic,
park2025mitigating,lee2026enhancing,lee2025video,ranasinghe2024understanding}.
The view associated with the selected answer is denoted by
$H_l^{\mathrm{final}}$.
Repeating this process for all $L$ candidate groups yields
\begin{equation}
    \mathcal{H}^{\mathrm{final}}
    =
    \{H_l^{\mathrm{final}}\}_{l=1}^{L},
    \qquad
    H_l^{\mathrm{final}}
    \in
    \mathcal{H}_l^{\mathrm{candi}}.
\end{equation}
Our results empirically demonstrate the effectiveness of this selection
strategy, as shown in~\Tref{tab:ablation_llm_design}.

\paragraph{Verification} 
Following the final view selection stage, we perform verification by comparing the answers from the initial and final views. 
We prioritize the final view, particularly when the initial view fails to provide a definitive answer.
Recent advancements in VLMs have led to behaviors in which models state responses, such as ``I don't know''~\cite{zhang2024r}, when the visual cues are insufficient.
Our pipeline mitigates this issue by identifying and selecting the specific view that provides the necessary visual cues, thereby improving overall performance.


\paragraph{Number of views by downstream task}
The final view set
$\mathcal{H}^{\mathrm{final}}
=\{H_l^{\mathrm{final}}\}_{l=1}^{L}$
contains one selected view from each local candidate group.
For both embodied question answering (EQA) and grounding, all $L$ final
views are provided to the VLM for reasoning.

For EQA, all final views are directly used for question answering, such that
$\mathcal{H}^{\mathrm{task}}=\mathcal{H}^{\mathrm{final}}$ and $L'=L$.
For grounding, the VLM additionally returns the image indices associated
with the referred targets.
The corresponding final views form the task-view set
\begin{equation}
    \mathcal{H}^{\mathrm{task}}
    =
    \{H_k^{\mathrm{task}}\}_{k=1}^{L'}
    \subseteq
    \mathcal{H}^{\mathrm{final}}.
\end{equation}
We set $L'=1$ for single-target object
localization~\cite{chen2020scanrefer}, whereas $L'$ is determined dynamically
with $L'\le L$ for multi-target object
localization~\cite{zhang2023multi3drefer}.

\subsection{Downstream Tasks}\label{subsec:downstream_task}

%
\paragraph{Episodic embodied question-answering (EM-EQA)}
We render the task poses into RGB images and provide the resulting images together with the query $\mathcal{Q}$ to a VLM~\cite{openai2023gpt4v}.
Compared with methods that rely only on pre-recorded views~\cite{gu2024conceptgraphs,majumdar2024openeqa,3d-mem}, our VLM-as-Judge with novel-view
synthesis provides visual evidence from viewpoints that are unavailable in the original episodic memory.

\paragraph{3D visual grounding}
For 3D visual grounding, the selected task poses provide spatial constraints for localizing the referred targets in the Gaussian scene.
We assume shared camera intrinsics, image bounds, and near/far clipping planes across all viewpoints, and omit these fixed parameters from the frustum notation.

Let $\mathcal{F}(H_k^{\mathrm{task}})$ denote the viewing frustum of the $k$-th task pose.
We retain the activated Gaussians whose 3D centers lie within the task-view frustums:
\begin{equation}
    \mathcal{G}^{\mathrm{frustum}}
    =
    \left\{
        G_j\in\mathcal{G}^{\mathrm{act}}
        \;\middle|\;
        \mathbf{x}_j
        \in
        \bigcup_{k=1}^{L'}
        \mathcal{F}(H_k^{\mathrm{task}})
    \right\},
\end{equation}
where $\mathbf{x}_j\in\Real^3$ denotes the 3D center of Gaussian $G_j$.
For single-target grounding, $\mathcal{G}^{\mathrm{frustum}}$ is the final predicted region.
For multi-target grounding, the frustum-restricted Gaussian subset associated with each task view is retained as a separate predicted instance candidate, while $\mathcal{G}^{\mathrm{frustum}}$ denotes their aggregate union.
\section{Experiments}
\label{sec:exp}
\nickname integrates 3D Gaussian Splatting (3DGS) and Vision-Language Models (VLMs), enabling selection of final viewpoints that directly correspond to the user's query.
The final viewpoints highlight the specific regions relevant to the user query, enabling embodied behavior such as embodied question-answering in~\Sref{subsec:eqa} or 3D grounding including 3D object localization (single or multiple targets with declarative query) and 3D referring segmentation (question query) in~\Sref{subsec:obj_loc}.
The necessity of our pipeline is shown by ablation studies in~\Sref{subsec:ablation}.

\begin{table}[t]
    \centering
    \caption{\textbf{Evaluation on EM-EQA.}
    We measure the semantic equivalent using LLM-Match and frame efficiency using the average number of frames.
    Methods marked with * and Human score are reported from OpenEQA, while all other results are reproduced.
    We report OpenEQA~\cite{majumdar2024openeqa} results since the code is not publicly available.
    Average frames denote the number of final views are used to answer queries.
    $\dagger$ indicates that while novel views are utilized for the VLM-as-Judge, the actual number of input frames fed into the VLMs remains consistent with the reported average frames.
    }
    \resizebox{.7\linewidth}{!}{
        \begin{tabular}{lccc}
            \toprule
            \multirow{2}{*}[-2pt]{Method} & \multicolumn{2}{c}{LLM-Match ($\uparrow$)} &  \multirow{2}{*}[-2pt]{Average Frames} \\ \cmidrule(lr){2-3}
            & Closed Model~\cite{hurst2024gpt4o} & Open Model~\cite{bai2025qwen3} &   \\
            \midrule
             BlindLLM~\cite{openai2023gpt4} & 34.8 & 29.4 & 0 \\
             Frame Captions  & 24.1 & 31.7 & 0 \\
             CG Captions\footnotemark[1]~\cite{gu2024conceptgraphs}  & 36.5 & - & 0 \\
             SVM Captions\footnotemark[1]~\cite{majumdar2024openeqa}  & 38.9 & - & 0 \\
             Multi-Frame  & 49.1 & 48.2 & 3.0 \\
             3D-Mem~\cite{3d-mem}  & 54.6 & 50.8 & 2.7  \\
             \cellcolor{oursblue}\nickname (ours)  & \cellcolor{oursblue}\best{57.8} & \cellcolor{oursblue}\best{51.6} & \cellcolor{oursblue}\best{2.6}\footnotemark[2] \\
             \midrule
             Human & \multicolumn{2}{c}{86.8} & Full \\
             \bottomrule
        \end{tabular}
    }\vspace{4mm}
    \label{tab:em-eqa}
    \centering
    \caption{
    \textbf{3D grounding results.}
    We evaluate recent studies, such as Dr. Splat, using both category-level and sentence-based queries.
    ReferSplat is tested with sentence-based queries, as it is specifically designed to interpret detailed textual descriptions.
    We evaluate our method on 3D object localization using ScanRefer and Multi3DRefer.
    }
    \resizebox{0.99\linewidth}{!}{
        \begin{tabular}{lccccccccc}
            \toprule
            \multirow{2}{*}[-2pt]{Method} & \multicolumn{3}{c}{ScanRefer~\cite{chen2020scanrefer}} & \multicolumn{3}{c}{Multi3DRefer~\cite{zhang2023multi3drefer}} & \multicolumn{3}{c}{3D Referring Segmentation} \\
            \cmidrule(lr){2-4} \cmidrule(lr){5-7} \cmidrule(lr){8-10}
            & 3D mIoU $\uparrow$ & Acc@8 $\uparrow$ & Acc@10 $\uparrow$ & 3D mIoU $\uparrow$ & Acc@8 $\uparrow$ & Acc@10 $\uparrow$ & 3D mIoU $\uparrow$ & Acc@8 $\uparrow$ & Acc@10 $\uparrow$ \\
            \midrule
             Dr. Splat (category)~\cite{drsplat} & 8.73 & 34.53 & 28.53 & 4.76 & 18.75 & 13.94 & 10.03 & 41.71 & 33.86   \\
             Dr. Splat (sentence)~\cite{drsplat} & 9.44 & 34.28 & 29.35 & 4.19 & 15.70 & 13.04 & 10.56 & \textbf{45.21} & 43.21 \\
             ReferSplat~\cite{refersplat} & 3.14& 10.68 & 7.32 & 1.27 & 2.19 & 1.40 & 2.34 & 2.04 & 0.00 \\
             \midrule
             \cellcolor{oursblue}\nickname (ours) & \cellcolor{oursblue}\textbf{11.12} & \cellcolor{oursblue}\textbf{35.84} & \cellcolor{oursblue}\textbf{32.75} & \cellcolor{oursblue}\textbf{6.44} & \cellcolor{oursblue}\textbf{31.24} & \cellcolor{oursblue}\textbf{25.27} & \cellcolor{oursblue}\textbf{12.46} & \cellcolor{oursblue}45.14 & \cellcolor{oursblue}\textbf{45.14}\\
             \bottomrule
        \end{tabular}
    }
    \label{tab:fine_grained}
\end{table}

\paragraph{Dataset}
We evaluate \nickname on two datasets that assess its capabilities in embodied reasoning and 3D scene understanding. 
To measure embodied reasoning capability, we use the evaluation dataset released by OpenEQA~\cite{majumdar2024openeqa}. 
Specifically, we focus on the Episodic-Memory Embodied Question Answering (EM-EQA) task, which reflects practical scenarios, \textit{e.g.}, smart glasses that rely on the history of past observations to assist users. 
OpenEQA question-answer pairs are constructed from Habitat-Matterport 3D~\cite{ramakrishnan2habitat} and ScanNet~\cite{dai2017scannet, scannet200}, covering seven types of questions with over 1,600 queries in about 180 environments.
 
In addition, we also evaluate 3D grounding with complex linguistic user queries, such as 3D object localization~\cite{chen2020scanrefer, zhang2023multi3drefer} and 3D referring segmentation~\cite{dai2017scannet, majumdar2024openeqa}.
In 3D object localization, we evaluate our model under two configurations: a single-target setting~\cite{chen2020scanrefer}, where each query corresponds to exactly one object, and a multi-target setting~\cite{zhang2023multi3drefer}, where a query can refer to multiple objects.
Since existing benchmarks predominantly use point clouds, we curate a 3DGS-based evaluation.
For the 3D object localization task, we utilize ScanRefer~\cite{chen2020scanrefer} validation set (141 scenes \& 9,508 queries), \textit{e.g.}, this is   the bed with blue sheets near the desk and Multi3DRefer~\cite{zhang2023multi3drefer} multi-target validation set (133 scenes \& 2,757 queries), \textit{e.g.}, this is a black chair facing the door.
To address more diverse formats of user queries such as questions, we manually generate 49 spatial queries (\textit{e.g.}, ``Which bed is closest to the window?'') that cover diverse attribute-based and spatial relationships.
Each scene contains multiple instances from the same object category, requiring the model to distinguish between them using precise spatial cues and attributes.

\subsection{Embodied Question Answering (EQA)}
\label{subsec:eqa}
\paragraph{Comparisons}
We compare our method with six types of agents. BlindLLM~\cite{openai2023gpt4, yang2025qwen3} infers their answer without visual input and only with the user question. 
Frame Captions uses LLaVA-v1.5~\cite{liu2024llavaimproved} to caption 50 sampled frames, feeding captions and the question to LLM~\cite{openai2023gpt4, yang2025qwen3}. 
CG Captions and SVM Captions construct textual scene graphs from episodic memory using ConceptGraphs~\cite{gu2024conceptgraphs} and Sparse Voxel Map~\cite{majumdar2024openeqa}, respectively. 
Multi-Frame directly provides three frames through unified sampling along with the question.
3D-Mem~\cite{3d-mem} leverages object-centric multi-view Memory Snapshots to preserve spatial and semantic consistency while reducing redundant frames. 
Ours utilizes semantic 3DGS as an episodic memory integrated with VLM~\cite{hurst2024gpt4o, bai2025qwen3} interaction, enabling a compact, informative, and effective representation. 

We reproduce all results except for CG and SVM Captions, for which the code and prompts are not publicly available. 
For the 3D-Mem evaluation, we follow the extracted snapshots provided by the 3D-Mem authors, which contain a small omission (about 1\%), and therefore evaluate 1,623 questions for all reproductions. 
As 3D-Mem constructs its memory based on closed-set categories, we also use closed-set semantic categories in our method for a fair comparison.

\paragraph{Settings}
We follow the EM-EQA setting proposed in OpenEQA~\cite{majumdar2024openeqa}, where the agent answers a question solely from past visual observations without further exploration. 
Each agent receives episodic memories or sampled frames as input and generates an open-ended textual answer via its VLMs.
Performance is evaluated using the LLM-Match metric~\cite{majumdar2024openeqa} 
which measures semantic agreement between predicted and ground-truth answers by prompting a strong language model~\cite{openai2023gpt4} to judge whether the two answers are semantically equivalent. 
This metric better captures conceptual correctness than exact string matching, 
providing a more reliable assessment of reasoning and grounding performance in embodied settings.
We test our method with both of a closed-source model (GPT-4o)~\cite{hurst2024gpt4o} and an open-source model (Qwen3-VL-8B)~\cite{bai2025qwen3} for the reproducibility and accessibility.

\paragraph{Results}
\Tref{tab:em-eqa} reports the LLM-Match performance and average frames across different baselines. 
BlindLLM and Captions achieve relatively low scores indicating that relying solely on textual information is insufficient for embodied reasoning. 
Incorporating visual signals significantly boosts performance: Multi-Frame results highlight the importance of visual context for understanding and reasoning. 
3D-Mem further improves performance through its compact yet informative Snapshot Memory architecture. 
Finally, \nickname achieves the highest performance while maintaining a similar average number of frames, validating the effectiveness and efficiency of our semantic 3DGS representation and novel-view adjustment approach.
For the open-source model, we employ the Qwen3-VL-8B model. Although it shows a relatively constrained performance gap due to the smaller model capability compared to closed-source models, we still observe consistent improvements. 
We show this performance gap to widen significantly with more capable, larger-scale models in the supplementary material.

\begin{figure}[!t]
    \centering
    \includegraphics[width=1.0\linewidth]{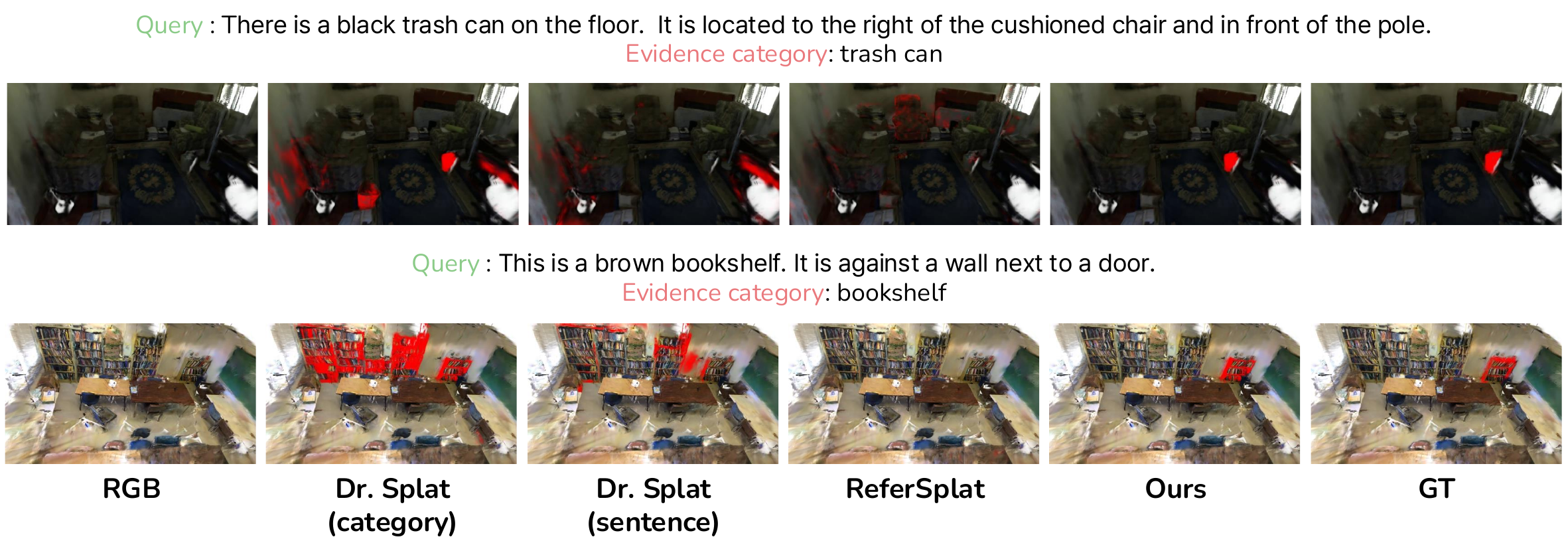}
    \vspace{-6mm}
    \caption{
        \textbf{
        Qualitative results of the 3D object localization (ScanRefer, single target).
        }
        Compared to the competing methods, ours more accurately identifies the fine-grained target locations referenced by the sentence query.
    }
    \label{fig:qual_scanrefer}
    \centering
    \includegraphics[width=1.0\linewidth]{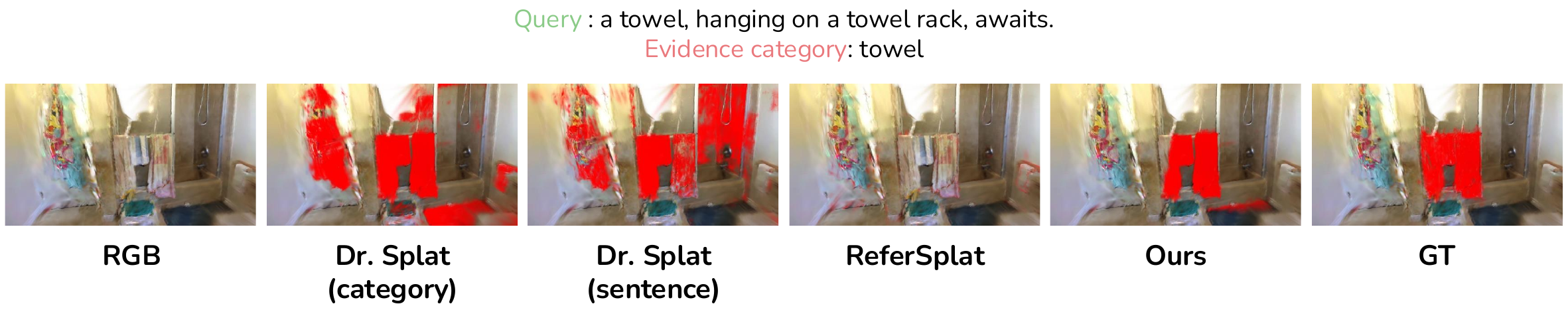}
    \vspace{-6mm}
    \caption{
        \textbf{
        Qualitative results of the 3D object localization (Multi3DRefer, multiple targets).
        }
        Compared to competing methods, ours achieves fine-grained localization even when a query references multiple target locations.
    }
    \label{fig:qual_multi3drefer}
    \centering
    \includegraphics[width=1.0\linewidth]{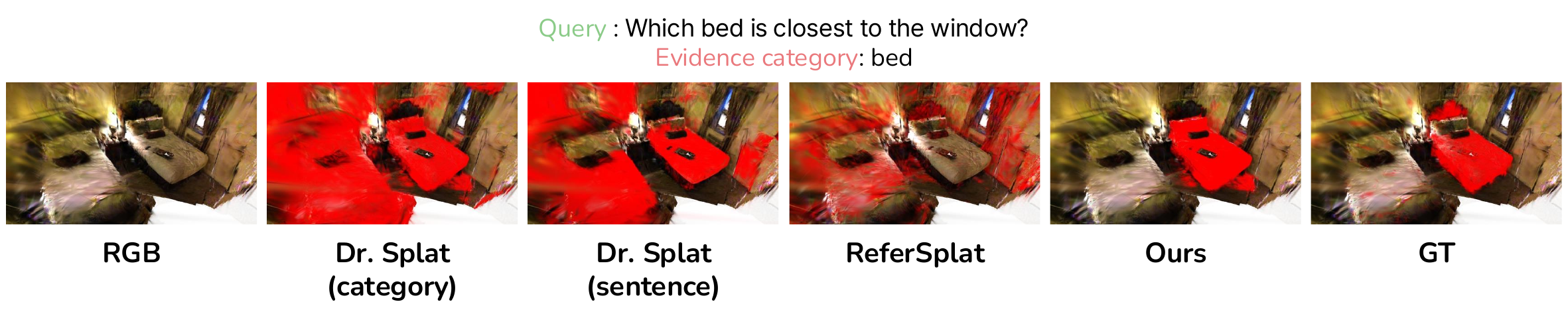}
    \vspace{-6mm}
    \caption{
        \textbf{
        Qualitative results of the 3D referring segmentation.
        }
        Our model can process queries in question formats.
    }
    \vspace{-4mm}
    \label{fig:qual_3d_loc}
\end{figure}

\subsection{3D Grounding}
\label{subsec:obj_loc}
\paragraph{Competing methods}
We compare our method against recent language-guided 3D Gaussian Splatting approaches: Dr. Splat~\cite{drsplat} and ReferSplat~\cite{refersplat}. 
Dr. Splat lifts multi-view CLIP~\cite{clip} image embeddings into the 3D Gaussian space for language-based querying. 
We optimize the model following the original paper's settings.
ReferSplat~\cite{refersplat} optimizes 3D Gaussians using fine-grained textual expressions for referring understanding.
Because ReferSplat requires paired textual annotations for supervision but lacks an official automated data generation pipeline, we manually constructed the necessary training data by using foundation models~\cite{yuan2024osprey,sam} to generate textual descriptions for the object masks.

\begin{figure}[!t]
    \centering
    \includegraphics[width=1.0\linewidth]{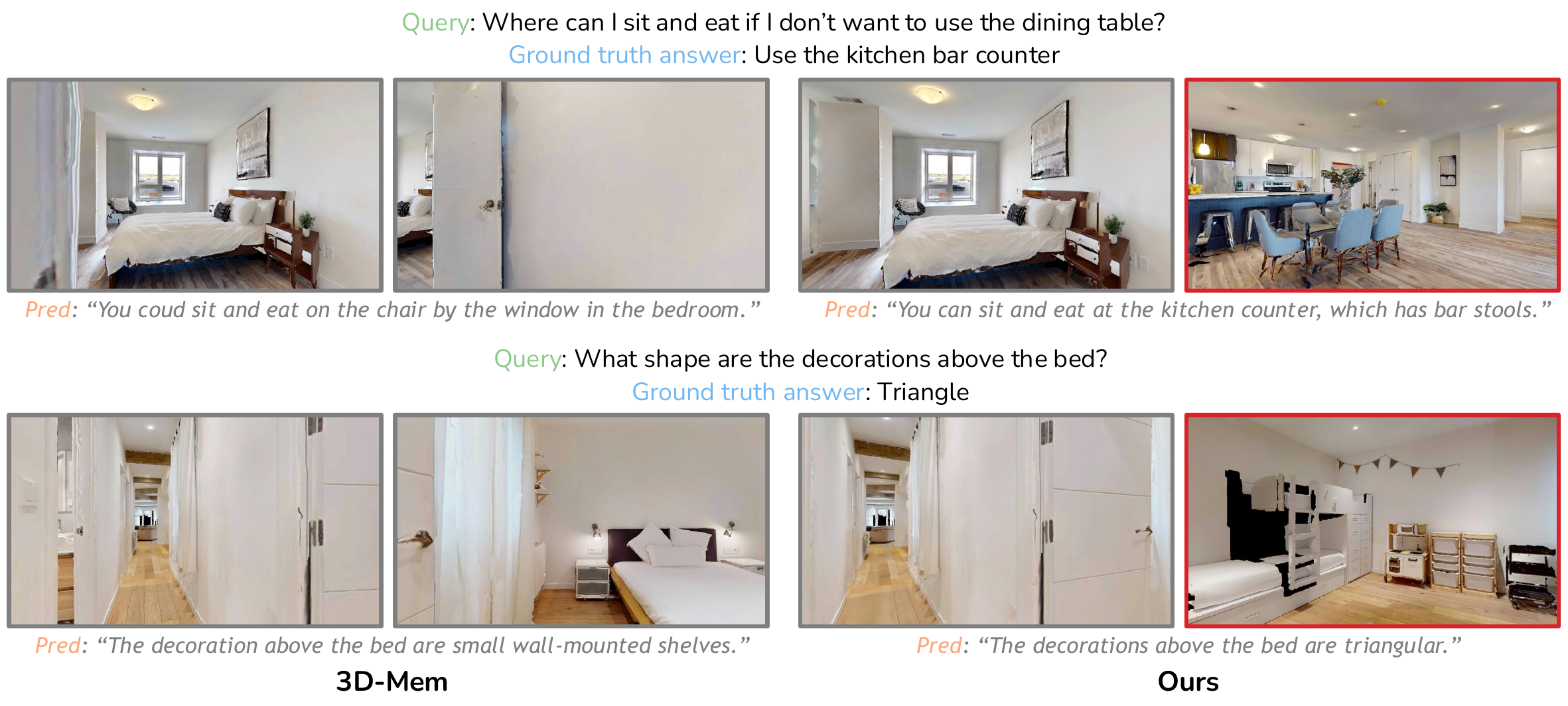}
    \vspace{-6mm}
    \caption{\textbf{Initial view selection.}
    Compared to 3D-Mem, our visibility-based policy can select more informative views by accounting for the actual visibility of target objects, rather than relying solely on the existence of categories within a given view.
    }
    \label{fig:initial_view}
\end{figure}

\begin{figure}[!t]
    \centering
    \includegraphics[width=0.97\linewidth]{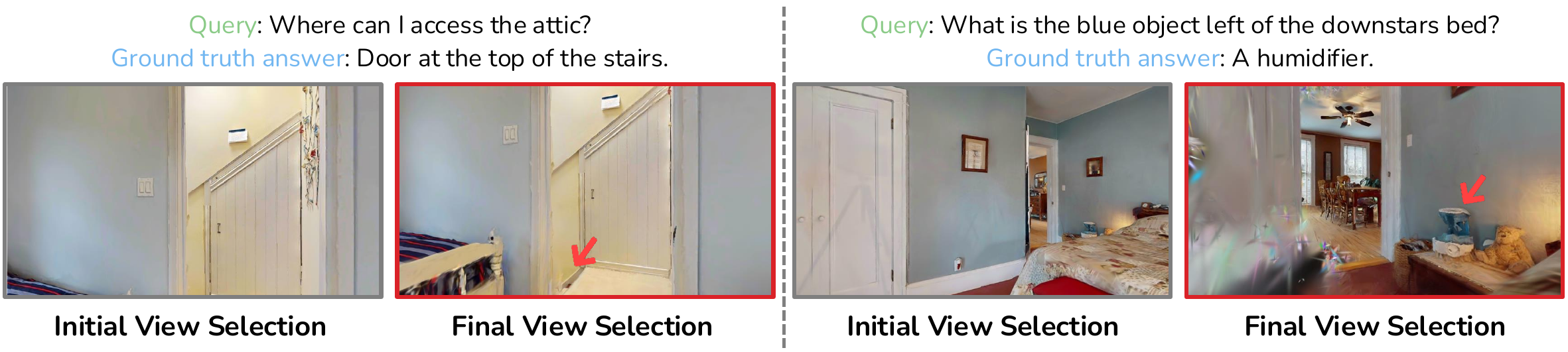}
    \caption{\textbf{Final view selection.}
    We apply novel view synthesis to refine initial views, leveraging 3DGS capability.
    Our results show that even when the initial views lack sufficient information to answer the question, the novel-view adjustment process can recover the necessary viewpoints, enabling more accurate question-aligned answers.
    }
    \label{fig:final_view}
    \vspace{-4mm}
\end{figure}

\paragraph{Settings}
We measure the 3D mean Intersection over Union (3D mIoU) between the activated Gaussians and the ground-truth (GT) Gaussians.
Since the scene is represented by Gaussians, we compute the mIoU based on the volumetric overlap induced by the Gaussian scales and opacities, following~\cite{drsplat}.

For 3D object localization, we distill the instance segmentation GT from ScanRefer~\cite{chen2020scanrefer} and Multi3DRefer~\cite{zhang2023multi3drefer} onto the 3DGS. 
For 3D referring segmentation, we manually annotate the target instance corresponding to each query, as existing benchmarks lack grounding annotations for question-form queries.

The ground-truth 3D Gaussian labels are distilled by computing the Mahalanobis distance between the ground-truth point clouds from the dateset~\cite{dai2017scannet} and 3D Gaussians, following the procedure in~\cite{drsplat}.
Given a user query (sentence or question), each model identifies the corresponding activations over all 3D Gaussians in the scene.
A higher mIoU indicates better grounding accuracy, meaning that the activated Gaussians more precisely align with the target region referenced by the user query.
For details of our evaluation dataset construction, please refer to the supplementary material.

\paragraph{Results}
\Tref{tab:fine_grained} reports 3D mIoU and Acc@k scores. 
Acc@k denotes the percentage of predictions whose IoU with the ground truth exceeds k\%.
We normalize the computed Gaussian volumes by the 90$\%$ value and clip their range between 0 and 1, which helps suppress excessive floating Gaussians that appear in vanilla 3DGS~\cite{fan2024lightgaussian}.
While ReferSplat is designed to interpret detailed textual descriptions, its rendering-based optimization is poorly aligned with direct 3D search, resulting in lower mIoU. 
In contrast, \nickname more reliably identifies the question-related 3D regions thanks to the combination of directly registered semantic 3DGS and VLM-based reasoning.

In Figs.~\ref{fig:qual_scanrefer}, \ref{fig:qual_multi3drefer}, and \ref{fig:qual_3d_loc}, we observe that our method aligns more closely with the ground truth, demonstrating stronger fine-grained grounding capability. 
This capability is particularly important for embodied exploration and navigation tasks, where the system must precisely reflect user queries.
Dr. Splat is designed to find objects without distinguishing instances, which limits its ability to localize fine-grained target regions (category) and, consequently, makes it struggle to directly interpret complex language queries (sentence).
ReferSplat is designed to respond to handle complex language queries, but is rendering-based optimization is not compatible with direct 3D search, leading to limited 3D localization.

\subsection{Ablation Study}
\label{subsec:ablation}
We use 184 embodied question answering (EQA) questions for initial design choices, and evaluate verification stage on the full set of 1,623 questions.
All ablation studies are conducted using closed-source models~\cite{hurst2024gpt4o, openai2023gpt4v, singh2025openaigpt5}.

\begin{wraptable}{r}{0.45\linewidth}
    \vspace{-12mm}
    \centering 
    \caption{\textbf{Ablation on initial view selection strategy (184 questions).}
    Because 3DGS is volumetric, we also compute volumetric visibility scores in addition to the non-volumetric version.
    }
    \resizebox{.97\linewidth}{!}{
        \begin{tabular}{lcc}
             \toprule
             Method & LLM-Match ($\uparrow$)  & Average Frames \\
             \midrule
             3D-Mem~\cite{3d-mem} & 45.4 & 3.1 \\
             Volume score & 47.8 & 2.7 \\
             \cellcolor{oursblue}Visibility score (ours) & \cellcolor{oursblue}\best{48.2} & \cellcolor{oursblue}2.7  \\
             \bottomrule
        \end{tabular}    
    }
    \label{tab:ablation_initial_view}
    \vspace{-4mm}
\end{wraptable}

\paragraph{Score function for initial view}
We compare two score functions for selecting initial views: a volume score, computed by replacing the Gaussian count with
\begin{equation}
    \text{Volume}(G_j)=s^x_js^y_js^z_j \alpha_j,
\end{equation}
where $s$ denotes the scale values of each axis and $\alpha$ indicates the opacity of each Gaussian.
The volume-based formulation performs worse because large outlier Gaussians can disproportionately dominate the score, leading to strong bias and unreliable view selection.
In contrast, the visibility score effectively acts as a proxy for object surface coverage, independent of Gaussian scale or training dynamics. 
This yields more stable visibility estimates and produces noticeably better LLM-Match performance, as shown in~\Tref{tab:ablation_initial_view}.

\paragraph{View selection pipeline} 
As shown in~\Tref{tab:ablation_view-selection}, the initial view selection alone already improves performance over 3D-Mem, demonstrating the effectiveness of our visibility score. 
Figure~\ref{fig:initial_view} presents snapshots of 3D-Mem and our initial view selection. We observe that the visibility scores successfully guide the selection of more informative views.
With the final view selection, integrated with novel views and the verification stage, performance further increases, indicating that 3DGS helps uncover more informative viewpoints. 
\begin{wraptable}{r}{0.45\linewidth}
    \vspace{-12mm}
    \centering
    \caption{\textbf{Ablation study on final view selection (184 questions).}
    We demonstrate the necessity of each component in our method.
    }
    \resizebox{.97\linewidth}{!}{
        \begin{tabular}{lcc}
             \toprule
             Method & LLM-Match ($\uparrow$) & Average Frames \\
             \midrule
             3D-Mem~\cite{3d-mem} & 45.4 & 3.1 \\
             Initial view selection & 48.2 & 2.7 \\
             \cellcolor{oursblue}Final view selection (ours) & \cellcolor{oursblue}\best{50.5} & \cellcolor{oursblue}2.7\footnotemark[2] \\
             \bottomrule
        \end{tabular}    
    }
    \label{tab:ablation_view-selection}
    \vspace{-7mm}
\end{wraptable}
As shown in \Fref{fig:final_view}, final views better capture evidence relevant to the question, which in turn supports VLMs in generating more accurate answers.
$\dagger$ indicates that while novel views are utilized during the view adjustment stage, the actual number of input frames fed into the VLMs remains consistent with the reported average frames.

\begin{wraptable}{r}{0.45\linewidth}
    \vspace{-12mm}
    \centering
    \caption{\textbf{Ablation study on VLM-as-Judge design (184 questions).} 
    }
        \resizebox{0.97\linewidth}{!}{
            \begin{tabular}{lcc}
                \toprule
                Method & LLM-Match ($\uparrow$) & Average Frames \\
                \midrule
                3D-Mem~\cite{3d-mem} & 45.4 & 3.1 \\
                Image & 47.1 & 2.7\footnotemark[2] \\
                Image + Text & 46.9 & 2.7\footnotemark[2] \\
                \cellcolor{oursblue}Text (ours) & \cellcolor{oursblue}\best{50.5} & \cellcolor{oursblue}2.7\footnotemark[2] \\
                \bottomrule
            \end{tabular}    
        }
    \label{tab:ablation_llm_design}
    \vspace{-7mm}
\end{wraptable}

\paragraph{VLM-as-Judge design}
For the final view selection, we perform a VLM-as-Judge process for novel-view adjustment.
To evaluate the informativeness of each candidate view, we rely solely on text-based descriptions and feed them into LLMs~\cite{ranasinghe2024understanding}.
As demonstrated in~\Tref{tab:ablation_llm_design}, our design choice empirically outperforms VLMs that incorporate visual information, while offering a more computationally efficient and lightweight alternative.

\begin{wraptable}{r}{0.45\linewidth}
    \vspace{-12mm}
    \centering
    \caption{\textbf{Ablation study on verification process (1,623 questions).} 
    }
    \resizebox{.97\linewidth}{!}{
        \begin{tabular}{lcc}
             \toprule
             Method & LLM-Match ($\uparrow$) & Average Frames \\
             \midrule
             Final view & 54.5 & 2.6\footnotemark[2] \\
             \cellcolor{oursblue}Final view + verification (ours) & \cellcolor{oursblue}\best{57.8} & \cellcolor{oursblue}2.6\footnotemark[2] \\
             \bottomrule
        \end{tabular}    
    }
    \label{tab:ablation_verification}
    \vspace{-7mm}
\end{wraptable}

\paragraph{Verification process}
In the verification stage, we compare initial views with final candidate views generated through novel-view adjustment.
A final viewpoint is selected only if it contains previously missing information.
As shown in ~\Tref{tab:ablation_verification}, our approach achieves improved performance compared to the baseline without the verification step.


\section{Conclusion}
\label{sec:conclusion}



We introduce \nickname, a framework utilizing 3D Gaussians for compact episodic memories to enable VLM reasoning and 3D grounding. By integrating semantic 3DGS, visibility-based view selection, and novel view synthesis, \nickname retrieves and refines query-relevant perspectives, overcoming the occlusions and detail loss typical of fixed-view systems.
While currently designed for passive episodic-memory scenarios rather than active robotic navigation, extending \nickname to active, long-horizon exploration with sequential decision-making remains a promising future avenue.


\vfill
{\scriptsize\paragraph{Acknowledgements}
This work was supported by Institute of Information \& Communications Technology Planning \& Evaluation (IITP) grant (No. RS-2020-II200004, Development of Previsional Intelligence based on Long-term Visual Memory Network (25\%); No. RS-2026-25518317, Development of AI memory mechanism that reflects human cognitive principles (25\%); No.RS-2019-II191906, Artificial Intelligence Graduate School Program(POSTECH) (25\%)), the National Research Foundation of Korea(NRF) grant (No. RS-2024-00453301 (25\%)), and the InnoCORE program (N10250156) funded by the Korea government (MSIT).
\par}

%
%
\bibliographystyle{splncs04}
\bibliography{main}

\clearpage
\appendix
\setcounter{figure}{0}
\setcounter{table}{0}
\setcounter{section}{0}
\setcounter{equation}{0}
\renewcommand{\thefigure}{S\arabic{figure}}
\renewcommand{\thetable}{S\arabic{table}}
\renewcommand{\thesection}{\Alph{section}}
\renewcommand{\theequation}{S\arabic{equation}}
\renewcommand{\theHequation}{supp.\arabic{equation}}
\renewcommand{\theHsection}{supp.\Alph{section}}
\renewcommand{\theHfigure}{supp.\arabic{figure}}
\renewcommand{\theHtable}{supp.\arabic{table}}
\begin{center}
{\Large\bfseries Supplementary Material}
\end{center}
\section*{Contents}
 \begin{itemize}
    \item \Sref{supp:implementation}. Implementation Details.
    \item \Sref{supp:exp_setup}. Experiment Setup.
    \item \Sref{supp:benchmark}. 3D Referring Segmentation Dataset Curation.
    \item \Sref{supp:results}. Additional Results.
    \item \Sref{supp:discussion}. Limitation and Discussion.
\end{itemize}
\section{Implementation Details}
\label{supp:implementation}
Our pipeline consists of language-embedded Gaussian construction, initial view selection in pre-captured images, and final-view selection via novel-view synthesis.

\subsection{Language-embedded Gaussian Construction}
Recent studies propose to utilize 3D Gaussian Splatting for open-vocabulary 3D scene understanding~\cite{langsplat,drsplat,wu2024opengaussian}. Commonly, these methods first run the original 3D Gaussian Splatting~\cite{3dgs}, and then allocate the semantic information on top of these 3D Gaussians.

\paragraph{3D Gaussian Splatting}
Following the original paper~\cite{3dgs}, we first optimize 3D Gaussian Splatting (3DGS) to minimize the rendering loss.
Additionally, we apply the depth regularization loss~\cite{kerbl2024hierarchical} to further improve reconstruction quality.
Each Gaussian $G_j$ is parameterized by its center position $\mathbf{x}_j=[x_j,y_j,z_j]^\top\in\Real^3$, along with a scaling matrix $S_j=\operatorname{diag}(s_j^x,s_j^y,s_j^z)$ and rotation $R_j$ that together define the 3D covariance matrix $\Sigma_j$.
Furthermore, it is defined by its opacity $\alpha_j\in\Real$ and spherical harmonics (SH) coefficients $\mathbf{sph}_j\in\Real^{(d+1)^2\times3}$ of degree $d$.

These rasterized and splatted Gaussians $\mathcal{G}$ are rendered to a 2D pixel color $\mathbf{c}$ at pixel $\mathbf{u}$ as:
\begin{equation}
    \mathbf{c}(\mathbf{u})
    =
    \sum_{j=1}^{M}
    T_j\widetilde{\alpha}_j\mathbf{c}_j,
    \qquad
    \widetilde{\alpha}_j
    =
    \alpha_j G_j^{2D}(\mathbf{u}),
\end{equation}
where the sum is evaluated over the contributing Gaussians in front-to-back order.
$T_j$ is the accumulated transmittance, $\widetilde{\alpha}_j$ is the effective opacity, and the color $\mathbf{c}_j$ is derived from the spherical harmonics coefficients $\mathbf{sph}_j$ given the camera viewing direction.
$G_j^{2D}$ is the 2D Gaussian function at pixel $\mathbf{u}$ obtained by splatting the Gaussian parameters~\cite{3dgs,zwicker2001ewa}.
The rendering contribution $T_j\widetilde{\alpha}_j$ is used in the visibility-based initial view selection described in~\Sref{subsec:best_view}.

The $N$ input images share an intrinsic matrix $\mathbf{K}\in\Real^{3\times3}$ and are associated with camera poses $\mathcal{H}=\{H_i\}_{i=1}^{N}$, where $H_i=[\mathbf{R}_i|\mathbf{t}_i]\in\Real^{3\times4}$.

\paragraph{Language-embedded 3D Gaussians}
Commonly, the language embedding from foundation models like CLIP~\cite{clip} is embedded on each 3D Gaussian, called language-embedded 3DGS. 
As described in the main paper, our method is compatible with both types of semantics: closed-set category labels~\cite{varghese2024yolov8} and open-set CLIP~\cite{clip} embeddings. 

For a fair comparison, we leverage the same semantics used in competing methods.
For the reasoning task (EM-EQA evaluation), 3D-Mem~\cite{3d-mem} utilizes the closed-set object detector~\cite{varghese2024yolov8}, so we also adopt these closed-set category semantics.
For the grounding task, previous studies~\cite{refersplat, drsplat} utilize CLIP vision embeddings~\cite{clip} or BERT~\cite{bert} text embeddings to perform open-set 3D segmentation.
We utilize Dr. Splat~\cite{drsplat} because it supports direct 3D search, a critical requirement for \nickname's 3D search and their rendering.
In contrast, ReferSplat~\cite{refersplat} is unsuitable for our framework since its learning objective is based on 2D rendering; conducting searches directly in 3D space leads to significant performance degradation.

\subsection{Initial View Selection}
Given a user query and the language-embedded 3D Gaussians, this step leverages the embedded semantic information to select initial camera views from the pre-recorded images to answer the question.

\paragraph{Evidence categories from user query}
The key objective of this step is to reduce the relevant search space in the 3D scene using language-embedded Gaussians. 
A core challenge with these research is that they primarily rely on simple, category-level text queries. 
Consequently, they struggle to capture the richer linguistic reasoning required by the complex, compositional instructions found in typical user queries.
To overcome this limitation and effectively query the language-embedded Gaussians, we first use Large Language Models (LLMs) to extract simplified ``evidence categories'' from the user query. 
These extracted evidence categories then serve as direct semantic queries to filter and retrieve the relevant 3D Gaussians.

Using the closed-set detector vocabulary $\mathcal{C}$ based on ScanNet200~\cite{scannet200} categories, we assign a category label $C_j$ to each Gaussian $G_j$.
Based on the categories $C_j$, we use Large-Language Models (LLMs) to identify the top-$k$ categories that are most relevant to the user query, which serve as our evidence objects.
Following the protocols of 3D-Mem~\cite{3d-mem}, we set $k=3$ for EM-EQA, utilizing GPT-4o~\cite{hurst2024gpt4o} and the same prefiltering prompt to obtain evidence objects. 
We set $k=1$ for all 3D grounding tasks, including Multi3DRefer~\cite{zhang2023multi3drefer}, to select the single category most relevant to the user query. 
For Multi3DRefer, we then retrieve all target instances corresponding to the selected category.

\paragraph{Clustering 3D Gaussians}
Once the evidence categories are extracted from the user query, we must identify the corresponding 3D regions. 
Crucially, simply retrieving a global pool of activated Gaussians is insufficient for effective view selection. For example, if a query involves a ``chair'', the activated Gaussians might correspond to multiple distinct chairs located in entirely different rooms. If the system were to select viewpoints based on the entire unclustered set, it would struggle to determine how many views are actually needed, likely resulting in a single viewpoint that misses objects outside its immediate field of view. Therefore, we perform spatial clustering on these activated Gaussians to separate them into distinct, instance-level groups. By isolating each object instance, we ensure the system can sample at least one optimal training view per instance, guaranteeing comprehensive visual coverage.

Formally, to construct these instance-level groups, we first determine the initial set of activated Gaussians, denoted as $\mathcal{G}^{\mathrm{act}}$. For closed-set semantics, we directly retrieve Gaussians whose categorical labels match the evidence categories; for open-set embeddings, we calculate the CLIP similarity between Gaussian features and the category text, applying a uniform threshold of 0.5 to filter the activations. After filtering, we apply our spatial clustering to $\mathcal{G}^{\mathrm{act}}$. Because spatial scales vary significantly across different environments, the specific clustering distance threshold is empirically determined per dataset (e.g., HM3D~\cite{ramakrishnan2habitat}, ScanNet~\cite{dai2017scannet}) and kept fixed within each respective dataset.

\begin{figure}[t]
    \centering
        \includegraphics[width=0.6\linewidth]{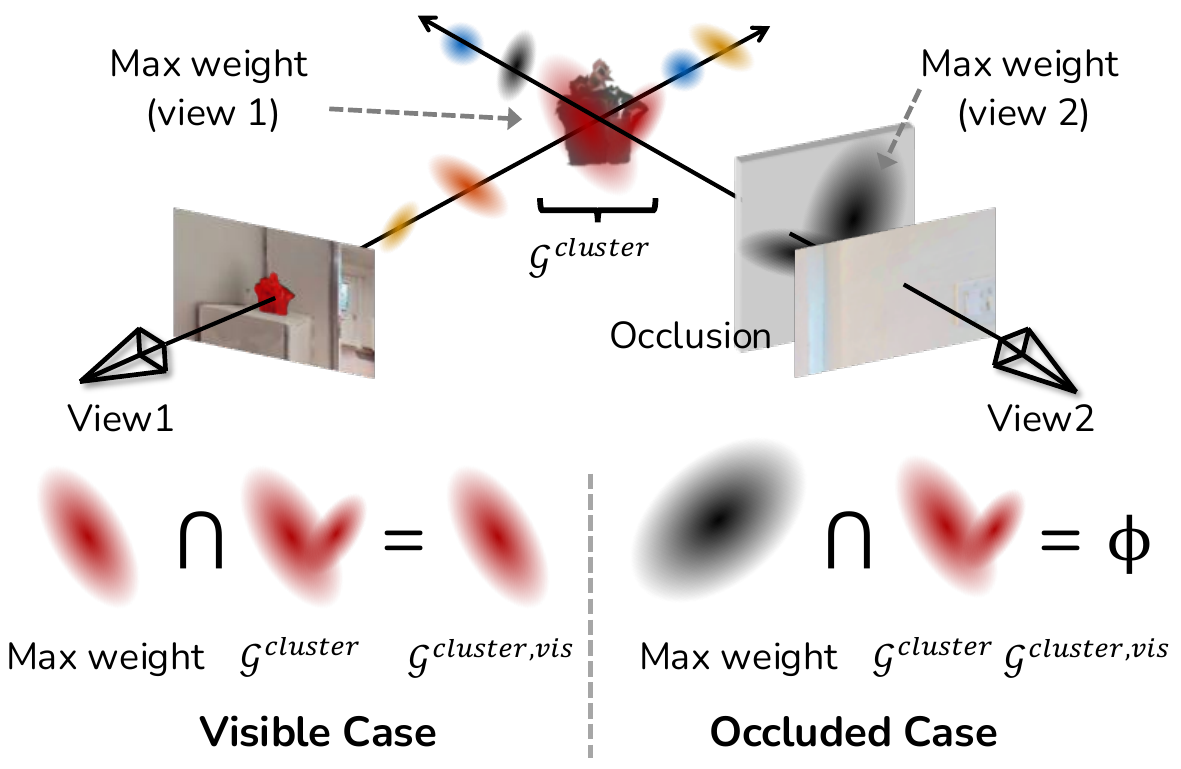}
    \caption{\textbf{Visibility Score.} 
    We evaluate visibility by checking whether Gaussians with the highest rendering weight belong to the target instance cluster. 
    If they align, the instance is considered visible; if occluded, the selected Gaussians originate from other regions, resulting in no overlap and a low visibility score.
    }
    \label{fig:visibility}
\end{figure}

\paragraph{Visibility score}
Rather than merely checking whether the target instance projects onto the 2D image plane, our visibility-based selection favors viewpoints from which it remains visible after accounting for occlusion by intervening 3D Gaussians, as illustrated in~\Fref{fig:visibility} and formalized in~\Sref{subsec:best_view}.

\subsection{Final View Selection with VLM-as-Judge}
After identifying the initial camera viewpoints, our framework refines them with a novel-view synthesis process.
Because the initial views may suffer from occlusions or suboptimal framing, we render multiple perturbed novel views using the 3DGS representation. 
We then employ a ``VLM-as-Judge'' pipeline to evaluate the visual evidence across a combined set of candidate viewpoints---consisting of both the initial and novel views, as visualized in~\Fref{fig:final_view_selection} of the main manuscript. 
For each of the $L$ local candidate groups, an LLM compares the textual answer candidates and selects one final view, yielding $\{H_l^{\mathrm{final}}\}_{l=1}^{L}$.

To implement this pipeline, we integrate specific Vision-Language Models (VLMs) and Large Language Models (LLMs). 
For the initial answer prediction stage, which uses a visual QA prompt to assess each candidate view, we employ GPT-4o~\cite{hurst2024gpt4o} as the closed-source VLM and Qwen3-VL-8B-Instruct~\cite{bai2025qwen3} as the open-source VLM, as illustrated in~\Fref{fig:final_view_selection} of the main paper and \Fref{fig:visual_qa}. 
For both the final view selection and the subsequent verification stages, we rely on text-based reasoning and use GPT-5-mini~\cite{singh2025openaigpt5} and Qwen3-8B~\cite{yang2025qwen3} as the LLMs.
For the EM-EQA stability test reported in~\Tref{tab:emeqa_stability}, we replace all VLM and LLM components with their corresponding alternative models.
Finally, to evaluate the LLM-Match metrics, we follow the protocols of previous methods~\cite{majumdar2024openeqa, 3d-mem} by using GPT-4.
Detailed prompts for view selection, verification, embodied question-answering, and 3D visual grounding are provided in \Fref{fig:view_selection}, \Fref{fig:verification}, \Fref{fig:eqa}, and \Fref{fig:grounding} respectively.

\section{Experiment Setup}\label{supp:exp_setup}

This section provides an extended description of the experiment setup, which could not be fully elaborated upon in the main paper due to space limitations.

\subsection{EM-EQA}
The original episodic-memory embodied question answering (EM-EQA) benchmark from OpenEQA~\cite{majumdar2024openeqa} contains 1,636 questions.
In our experiments, we follow the extracted snapshot information provided by the 3D-Mem~\cite{3d-mem} authors, which already includes a small omission.
As a result, we evaluate on 1,623 questions (approximately 1\% fewer), matching the setting used in their released data and results.
The omitted case corresponds to scene0648\_01 from the ScanNet~\cite{dai2017scannet} dataset.
We reproduce all previous methods reported in~\Tref{tab:em-eqa} in the main paper, except for CG Captions~\cite{gu2024conceptgraphs}, SVM Captions~\cite{majumdar2024openeqa}, and the Human results~\cite{majumdar2024openeqa}.
The first two cannot be reproduced because their official code is unavailable, while the Human score is an absolute reference collected over all 1,636 questions and thus cannot be re-evaluated under our experimental setting.

To compute the LLM-Match scores reported in the main paper, we use GPT-4-1106-preview, following the evaluation protocol of prior work~\cite{3d-mem}. For the supplementary experiments, we instead use GPT-4-Turbo, as GPT-4-1106-preview has since been deprecated.

\subsection{Ablation Study of EM-EQA}
We use the 184-question subset for design ablations and the full 1,623-question set for verification and stability evaluations.
This subset is identical to the active embodied question answering (A-EQA) split used in prior work~\cite{majumdar2024openeqa, 3d-mem}.

\subsection{3D Visual Grounding}

The original 3D localization~\cite{drsplat, wu2024opengaussian} task focuses on querying object categories from the semantic 3D Gaussians~\cite{drsplat}.
Our goal is to extend this task to a more fine-grained setting, where the system must handle complex natural-language user queries or questions, and identify the specific object instance among multiple candidates.
Dr. Splat~\cite{drsplat} is the representative work that shows such 3D localization task, and we follow the official implementation.
To address compositional natural language queries, ReferSplat~\cite{refersplat} introduces a referring segmentation framework for 3D Gaussian Splatting.
However, ReferSplat requires a dedicated referring-segmentation dataset for each 3D scene, and no official automatic data-generation pipeline is available.
To address this, we manually design a custom data generation pipeline using foundation models~\cite{yuan2024osprey, sam} to train ReferSplat.

\subsection{Evaluation of 3D Visual Grounding}

We present the evaluation protocols for single-target queries in ScanRefer~\cite{chen2020scanrefer} and 3D referring segmentation, and for multiple target queries in Multi3DRefer~\cite{zhang2023multi3drefer}.

\paragraph{Gaussian volume normalization}
For each Gaussian $G_j$, we compute its opacity-weighted volume as
\begin{equation}
    \text{Volume}(G_j) = s_j^x s_j^y s_j^z \alpha_j,
\end{equation}
where $s_j^x$, $s_j^y$, $s_j^z$, and $\alpha_j$ denote its scale values and opacity, respectively.
Following LightGaussian~\cite{fan2024lightgaussian}, we normalize $\text{Volume}(G_j)$ by the 90th-percentile value over all Gaussians and clip the result to the range $[0,1]$.
We denote the resulting normalized weight by $w_j$ and use it for all 3D visual grounding evaluations.

\paragraph{Evaluation on queries with single target}
We evaluate the performance on ScanRefer and 3D referring segmentation using a 3D mean Intersection-over-Union (mIoU) metric. 
Following Dr. Splat~\cite{drsplat}, the 3D mIoU is calculated based on the volumetric overlap between the predicted Gaussians and the ground truth target.
For each query, the 3D IoU between the predicted mask $\mathcal{M}_{pred}$ and the ground-truth mask $\mathcal{M}_{gt}$ is formulated as:
\begin{equation}
\text{3D IoU} = \frac{\sum_{j \in \mathcal{M}_{pred} \cap \mathcal{M}_{gt}} w_j}{\sum_{j \in \mathcal{M}_{pred} \cup \mathcal{M}_{gt}} w_j}
\end{equation}
where $j$ indexes the Gaussian elements within the respective intersection or union masks.
To compute the 3D mIoU, we average the 3D IoU scores over all evaluation queries.

\paragraph{Evaluation on queries with multiple targets}
To evaluate Multi3DRefer~\cite{zhang2023multi3drefer}, we adopt the instance-aware 3D mIoU.
Unlike single-target settings, multiple predicted instance candidates must be matched with multiple ground-truth targets.
In our pipeline, $\mathcal{G}^{\mathrm{act}}$ is first grouped into instance-level candidate clusters for viewpoint proposal.
The VLM then selects $L'$ task viewpoints from the final-view candidates.
The activated Gaussians retained within each selected task-view frustum constitute one predicted instance candidate.
These candidates form the final multi-target predictions, while their union is denoted by $\mathcal{G}^{\mathrm{frustum}}$.

Following the Hungarian assignment procedure of Multi3DRefer~\cite{zhang2023multi3drefer}, we compute pairwise volumetric IoUs and apply the Hungarian algorithm~\cite{kuhn1955hungarian} to obtain an optimal one-to-one correspondence between the predicted instance candidates and the ground-truth targets.
For our continuous instance-aware 3D mIoU, we do not apply an additional IoU threshold.
Each prediction--target pair assigned by the Hungarian algorithm is counted as a true positive (TP).
An unassigned predicted instance candidate is counted as a false positive (FP), and an unassigned ground-truth target as a false negative (FN).
Here, TP indicates the assignment status, whereas localization quality is captured by $\mathrm{IoU}_i$; thus, a poorly localized assigned pair is penalized by its low IoU.
The instance-aware 3D mIoU is formulated as:
\begin{equation}
    \text{Instance-aware 3D mIoU}
    =
    \frac{\sum_{i \in \mathrm{TP}} \mathrm{IoU}_i}
    {|\mathrm{TP}| + |\mathrm{FP}| + |\mathrm{FN}|}
\end{equation}
where $\mathrm{IoU}_i$ is the volumetric IoU between the $i$-th matched frustum-based predicted instance candidate and its corresponding ground-truth target, computed using the same Gaussian weighting scheme as in the single-target evaluation.

Under our direct-3D evaluation protocol, prior methods do not explicitly produce a variable number of instance-separated predictions.
By integrating a VLM, our method enables instance-aware multi-target grounding.

\begin{figure}[!t]
    \centering
    \includegraphics[width=1.0\linewidth]{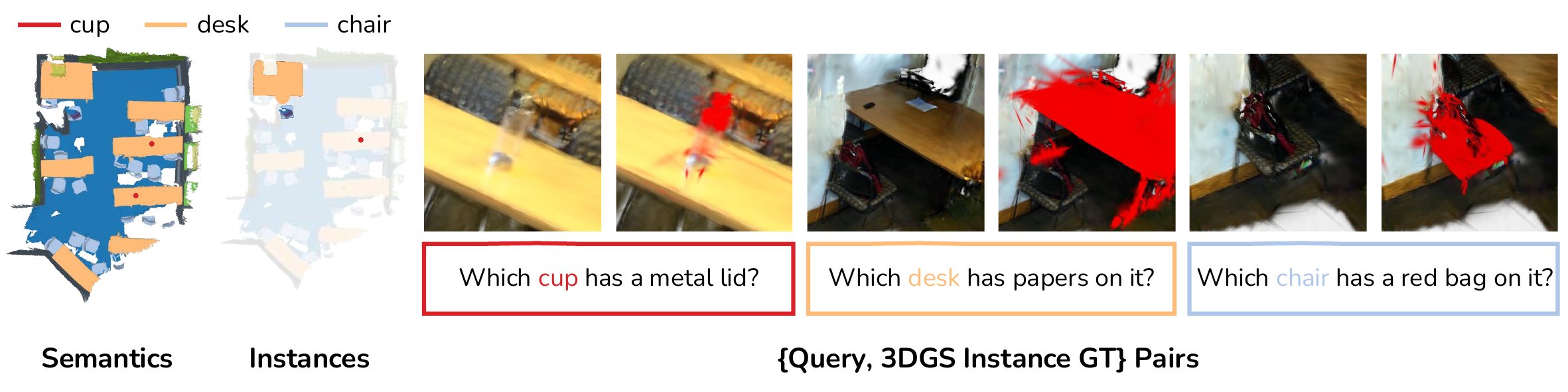}
    \caption{\textbf{3D referring segmentation dataset examples.}
    We curate a dataset consisting of user queries paired with 3DGS instance ground-truth (GT).
    }
    \label{fig:data_generation_pipeline}
\end{figure}

\section{3D Referring Segmentation Dataset Curation}\label{supp:benchmark}

We evaluate 3D grounding tasks such as 3D object localization (with declarative queries) and 3D referring segmentation (with interrogative queries).
Since no existing benchmarks provide paired queries and 3DGS instance ground-truth (GT), we construct our own by leveraging point-cloud-based semantic GT distillation, following the methodology described in~\cite{drsplat}.
The primary objective of our 3D grounding task is to distinguish a query-specific instance from multiple candidates belonging to the same semantic category, as illustrated in \Fref{fig:data_generation_pipeline}.
Previous language-embedded 3DGS methods~\cite{drsplat, wu2024opengaussian} have primarily focused on semantic segmentation rather than 3D grounding involving complex linguistic queries.
By integrating language-embedded 3DGS with VLM-based reasoning, \nickname effectively addresses these 3D grounding challenges.
To the best of our knowledge, this is the first-of-its-kind 3D grounding benchmark utilizing 3D Gaussian representations.

\subsection{Dataset Construction}

As our target task requires grounding objects directly within the 3D Gaussians, we distill the point-cloud based semantic GT into the Gaussian representation.
Specifically, we compute the Mahalanobis distance between each point and all Gaussians, accumulate these point-to-Gaussian assignments, and then apply an argmax operation to determine the instance label for each Gaussian.
The detailed GT distillation procedure follows the method described in~\cite{drsplat}.

\begin{table}[t]
    \centering
    \caption{\textbf{Ref-LERF vs. Our 3D referring segmentation.}
    Ours matches the scale of Ref-LERF in test query count but enables direct evaluation on 3D Gaussians in actual 3D space.
    While Ref-LERF identifies isolated targets, ours presents a more difficult challenge: differentiating the target instance from multiple similar instances.
    }
    \resizebox{.65\linewidth}{!}{
        \begin{tabular}{c|cccc}
            \toprule
            Method & \# Test Queries & GT Type & Metric & \# Candidate Instances \\
            \midrule
            Ref-LERF~\cite{refersplat}  & 59 & 2D Image & 2D mIoU & $1$ \\
            Ours  & 49 & 3D Gaussians & 3D mIoU & $\geq1$ \\
            \bottomrule
        \end{tabular}
    }
    \label{tab:data_stat}
\end{table}

\subsection{Statistics of Dataset}

We consider three settings: (1) ScanRefer~\cite{chen2020scanrefer} validation (single target), containing 9,508 single target queries from 141 scenes; (2) the Multi3DRefer~\cite{zhang2023multi3drefer} (multiple targets) validation split, containing 2,757 queries from 133 scenes; and (3) our 3D referring segmentation set (single target), containing 49 single target questions
from five scenes, where each target must be distinguished from same-category candidates.
We present what is, to the best of our knowledge, the first large-scale 3D object localization benchmark based on paired language and 3DGS instance data.

Although our 3D referring segmentation dataset is a curated subset of ScanNet, ours contains a comparable number of questions to Ref-LERF~\cite{refersplat}, while being constructed and evaluated directly in 3D space.
Furthermore, while both datasets assess spatial relationships, our setting requires distinguishing the target instance among multiple candidates, making it a more challenging task that is better aligned with practical embodied reasoning scenarios.
In contrast, Ref-LERF focuses on identifying a single target instance, as shown in~\Tref{tab:data_stat}.
While the original Ref-LERF dataset includes training and test splits, we report the size of the test split as we use these benchmarks exclusively for evaluation.

Beyond demonstrating strong performance on large-scale complex linguistic queries for 3D object localization, we further validated \nickname using an advanced interrogative (question-style) format on a curated subset in 3D referring segmentation.
As shown in~\Tref{tab:fine_grained} of the main paper, our method achieves consistent performance gains across both query formats, highlighting its robustness regardless of the linguistic structure.

\section{Additional Results}\label{supp:results}
In this section, we present additional experimental results that could not be included in the main paper due to space constraints.

\begin{table}[!t]
    \centering
    \begin{minipage}[t]{0.38\linewidth}
        \centering
        \caption{\textbf{Effect of 3DGS quality.}}
        \label{tab:3dgs_effect}
        \resizebox{\linewidth}{!}{
            \begin{tabular}{lc}
                \toprule
                Method & LLM-Match ($\uparrow$) \\
                \midrule
                3DGS w/o depth loss &  47.4 \\
                3DGS w/ depth loss (ours) & 51.8 \\
                \bottomrule
            \end{tabular}
        }
    \end{minipage}
    \hfill 
    \begin{minipage}[t]{0.28\linewidth}
        \centering
        \caption{\textbf{Effect of number of views.}}
        \label{tab:number_of_views}
        \resizebox{\linewidth}{!}{
            \begin{tabular}{ll}
                \toprule
                Avg. Frames & LLM-Match ($\uparrow$) \\
                \midrule
                1.3 & 40.9  \\
                1.7 & 41.7 \textcolor[HTML]{4285F4}{(+0.8)}\\
                2.7 (ours) & 51.8 \textcolor[HTML]{4285F4}{\textbf{(+10.1)}}\\
                4.6 & 57.2 \textcolor[HTML]{4285F4}{(+5.4)} \\
                \bottomrule
            \end{tabular}
        }
    \end{minipage}
    \hfill 
    \begin{minipage}[t]{0.28\linewidth}
        \centering
        \caption{\textbf{Sensitivity of perturbation $V$.}}
        \label{tab:perturbation}
        \resizebox{\linewidth}{!}{
            \begin{tabular}{ll}
                \toprule
                $V$ & LLM-Match ($\uparrow$) \\
                \midrule
                1 & 51.2 \\
                2 & 51.2 \textcolor[HTML]{4285F4}{(+0.0)} \\
                4 (ours) & 51.8 \textcolor[HTML]{4285F4}{\textbf{(+0.6)}} \\
                8 & 52.4 \textcolor[HTML]{4285F4}{(+0.6)} \\
                \bottomrule
            \end{tabular}
        }
    \end{minipage}
\end{table}

\begin{table}[!t]
    \centering
    \caption{\textbf{EM-EQA evaluation stability, three runs statistics (1,623 questions).}}
    \label{tab:emeqa_stability}
    \resizebox{0.99\linewidth}{!}{%
        \begin{tabular}{l|ccc|ccc|ccc}
            \toprule
            \multirow{2}{*}[-2pt]{Method} & \multicolumn{3}{c|}{Qwen3-VL-8B-Instruct} & \multicolumn{3}{c|}{Qwen3-VL-32B-Instruct}  & \multicolumn{3}{c}{Claude Sonnet 4.6} \\
            \cmidrule(lr){2-4} \cmidrule(lr){5-7} \cmidrule(lr){8-10}
            & 3D-Mem~\cite{3d-mem} & Ours & p-value & 3D-Mem~\cite{3d-mem} & Ours & p-value & 3D-Mem~\cite{3d-mem} & Ours & p-value \\ 
            \midrule
            LLM-Match ($\uparrow$) & 52.2 $\pm$ 1.21 & 52.7 $\pm$ 0.92 & 0.68658 & 56.4 $\pm$ 0.15  & 57.5 $\pm$  0.21 & 0.30821 & 53.2 $\pm$ 0.31 & 56.8 $\pm$ 0.12 & 0.00103 \\
            \bottomrule
        \end{tabular}%
    }
\end{table}

\begin{table}[!t]
    \centering
    \caption{\textbf{Inference time (seconds/query).} 
    }
    \resizebox{.70\linewidth}{!}{
        \begin{tabular}{ccccc}
             \toprule
             Clustering~~&~~Novel-view Rendering~~&~~VLM-as-Judge~~&~~Verification~~&~~\# Tokens~~\\
             \midrule
             0.93 & 31.68 & 41.20 & 0.40 & 6K \\
             \bottomrule
        \end{tabular}    
    }
    \label{tab:inference_time}
\end{table}

\subsection{Ablation Studies}

\paragraph{Effect of 3DGS quality (184 questions)}
Since our framework leverages images rendered from 3DGS for reasoning, the quality of the 3DGS representation can directly affect the reasoning capability. 
To evaluate this impact, we conduct experiments with and without depth loss supervision, as summarized in~\Tref{tab:3dgs_effect}. 
The results indicate that varying the depth loss supervision influences the outcomes, demonstrating high-quality 3DGS representations are crucial for robust reasoning.

\paragraph{Effect of the number of selected views (184 questions)}
We ablate the effect of the number of selected views in~\Tref{tab:number_of_views}, revealing significant performance gains with an average of only 2.7 frames. 
Furthermore, a separate experiment utilizing all novel views yields a LLM-Match score of 51.5. 
This underscores that our method efficiently isolates critical information, achieving favorable performance without requiring redundant views.

\begin{figure}[!t]
    \centering
    \includegraphics[width=1.0\linewidth]{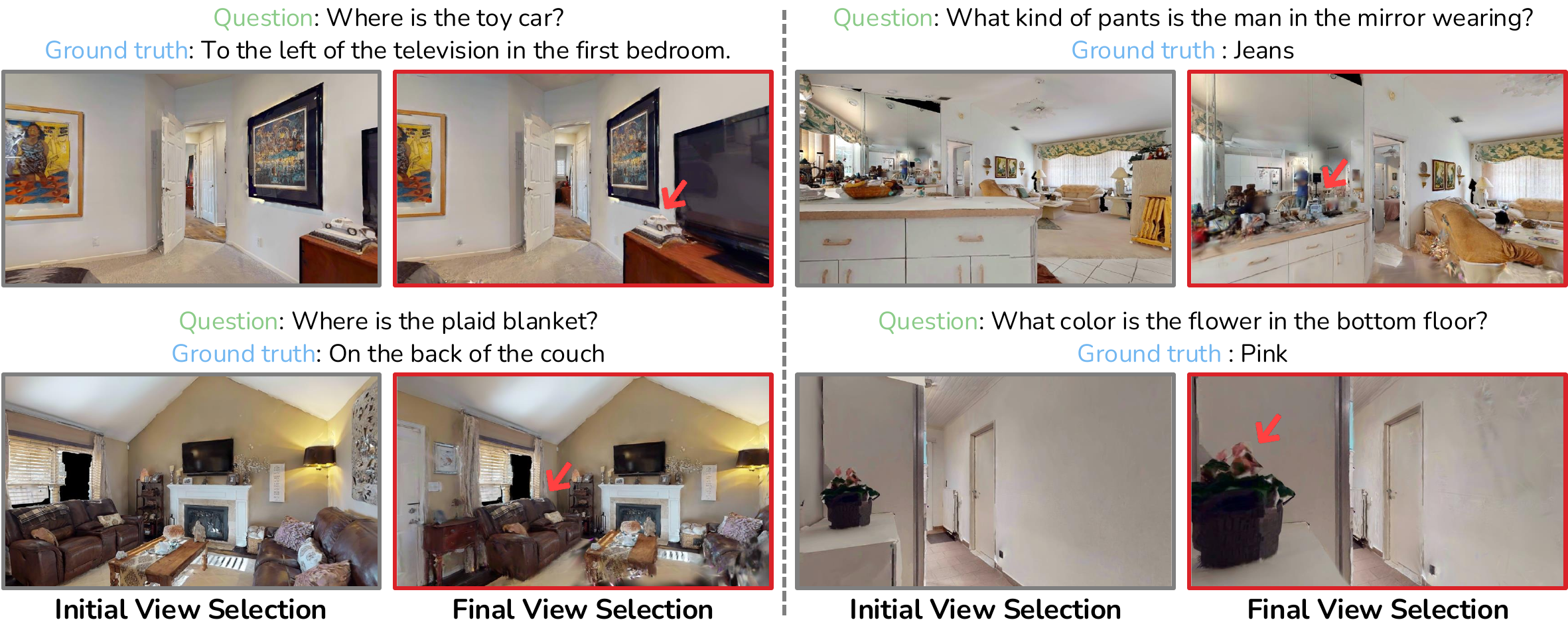}
    \vspace{-2mm}
    \caption{\textbf{Final view selection.}
    We apply novel view synthesis to refine the initially selected views, leveraging the unique capability of 3DGS.
    Our results show that when initial views are insufficient, the novel-view adjustment can recover informative viewpoints, yielding more accurate question-aligned answers.
    Empirically, VLMs tend to prefer evidence-centered or zoomed-in images compared to the initially selected view.
    }
    \label{fig:final_view_more}
\end{figure}

\paragraph{Sensitivity analysis of perturbation $V$ (184 questions)}
We evaluate the performance stability of our model by varying the perturbation parameter $V$, as summarized in~\Tref{tab:perturbation}. 
Note that our left and right views incorporate yaw-axis rotation. 
Setting $V=4$ (+2 views) yields a performance gain of +0.6. 
While further increasing $V$ to 8 (+4 views) provides an identical incremental gain of +0.6, it requires twice as many views. 
Therefore, $V=4$ serves as the optimal trade-off between performance and computational efficiency.

\paragraph{Performance stability of EM-EQA (1,623 questions)}
To demonstrate the performance stability of the EM-EQA evaluation, we conduct three independent runs and report the mean $\pm$ standard deviation along with a paired significance test at the question level. 
As shown in~\Tref{tab:emeqa_stability}, the performance gain for Qwen3-VL-8B is not statistically significant, which is bottlenecked by its limited reasoning capability. However, scaling to a stronger reasoning model (e.g., Claude Sonnet) yields a significant +3.6 gap ($p < 0.01$). 
This confirms that our method remains consistently effective and scales robustly with advanced reasoning capacity.




\begin{figure}[!t]
    \centering
    \includegraphics[width=1.0\linewidth]{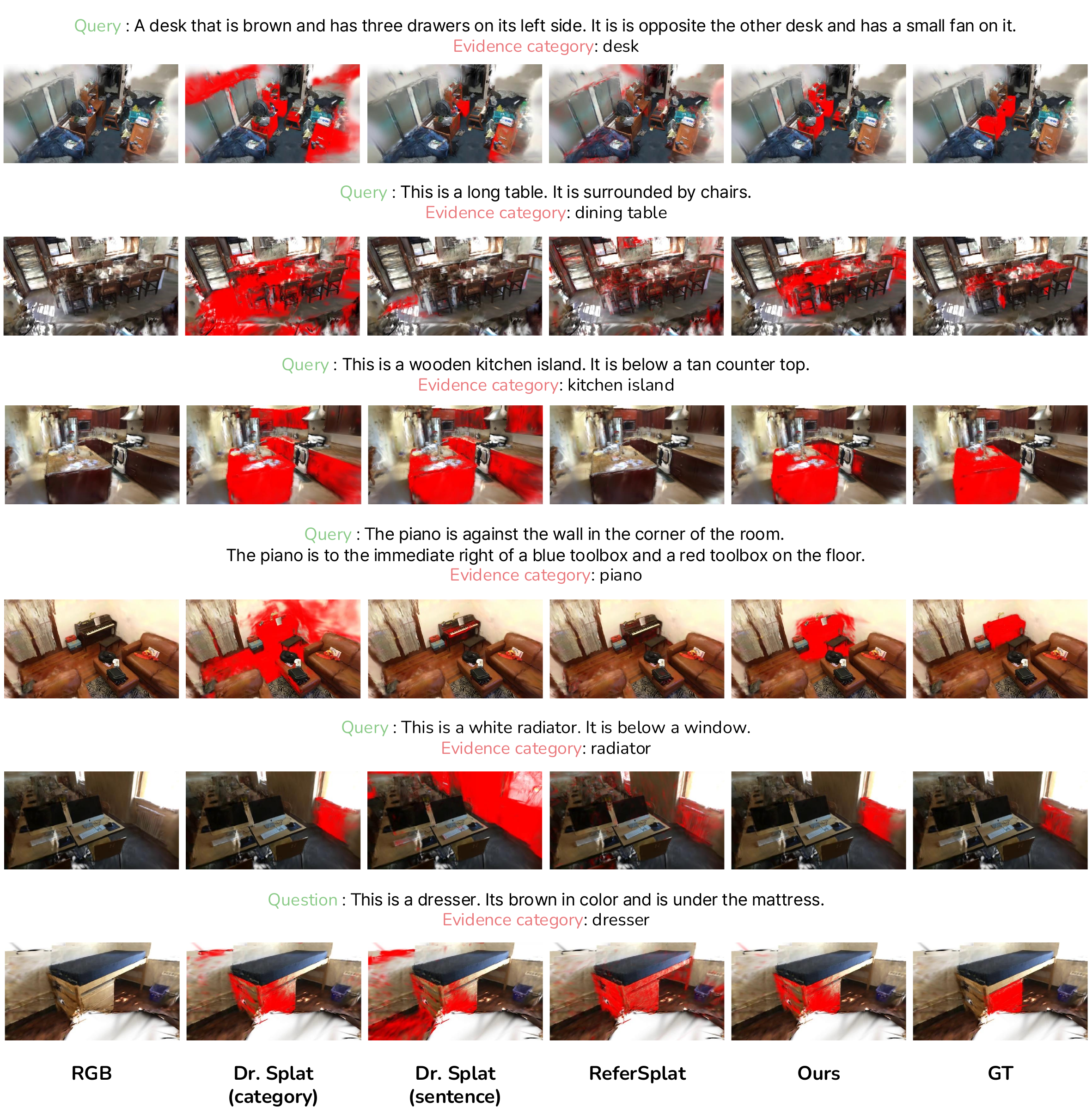}
    \caption{\textbf{Qualitative results of 3D object localization (ScanRefer, single target).}
    Compared to the competing methods, ours more accurately identifies the fine-grained target locations referenced by the user query.
    Dr. Splat is designed to find objects without distinguishing instances, which limits its ability to localize fine-grained target regions (category) and, consequently, makes it struggle to directly interpret complex language queries. 
    ReferSplat is designed to handle complex language queries, but its rendering-based optimization is not compatible with direct 3D search, leading to limited 3D localization capability.
    }
    \label{fig:scanrefer}
\end{figure}

\subsection{Inference Time}

We analyze the computational costs by breaking down the inference time per query in Table \ref{tab:inference_time}. 
Although our pipeline requires additional inference time, it eliminates the need for task-specific fine-tuning. 
By utilizing pre-trained 2D VLMs, our method maintains strong zero-shot generalization.
For the 3D scene optimization, one-time pre-processing steps per scene require about 30 minutes for 3DGS and 2 minutes for semantic Gaussians, respectively.
Once optimized, no retraining is needed for new queries.

\subsection{Final View Selection}

We provide additional qualitative results of novel view adjustment based on the final selected views, as shown in~\Fref{fig:final_view_more}.
Compared to the initially selected views, our method can capture more informative evidence relevant to the question, further demonstrating the necessity of using VLM-as-Judge in the novel-view adjustment stage.
In practice, VLMs often favor views that bring the key evidence closer to the center or zoom in on the relevant region.

\subsection{3D Object Localization}

We report additional 3D object localization results for the single target setting in~\Fref{fig:scanrefer}.
We evaluate Dr. Splat~\cite{drsplat} at both the category level and the sentence level.
At the category level, Dr. Splat can retrieve all relevant categories but fails to localize objects that are specific to the user query.
At the sentence level, it struggles to directly interpret complex language descriptions.
ReferSplat~\cite{refersplat} is designed to handle spatial language queries, but because it is optimized using a rendering-based loss, it cannot perform direct search in 3D.
In contrast, our method can finely distinguish the specific objects referred to by the user query.
The primary distinction lies in the query format: 3D object localization utilizes declarative sentence queries, whereas 3D referring segmentation employs diverse interrogative-style (question-format) queries.

\begin{table}[!t]  
    \centering
    \caption{
    \textbf{3D referring segmentation with novel view synthesis.}
    We evaluate recent studies, such as Dr. Splat, using both category-level and sentence-based queries.
    ReferSplat is tested with sentence-based queries, as it is specifically designed to interpret detailed textual descriptions.
    Our method follows the pipeline of \nickname.
    }
    \resizebox{0.5\linewidth}{!}{
        \begin{tabular}{lccc}
            \toprule
            \multirow{2}{*}[-2pt]{Method} & \multicolumn{3}{c}{3D Referring Segmentation} \\
            \cmidrule(lr){2-4}
            & 3D mIoU $\uparrow$ & Acc@8 $\uparrow$ & Acc@10 $\uparrow$ \\
            \midrule
             Dr. Splat (category)~\cite{drsplat} &  10.03 & 41.71 & 33.86   \\
             Dr. Splat (sentence)~\cite{drsplat} &  10.56 & 45.21 & 43.21 \\
             ReferSplat~\cite{refersplat} & 2.34 & 2.04 & 0.00 \\
             \midrule
             Ours & 12.46 & 45.14 & 45.14 \\
             Ours + novel-view & \textbf{12.87} & \textbf{45.71} & \textbf{45.71}  \\
             \bottomrule
        \end{tabular}
    }
    \label{tab:referring_novel_view}
\end{table}

\begin{figure}[!t]
    \centering
    \includegraphics[width=1.0\linewidth]{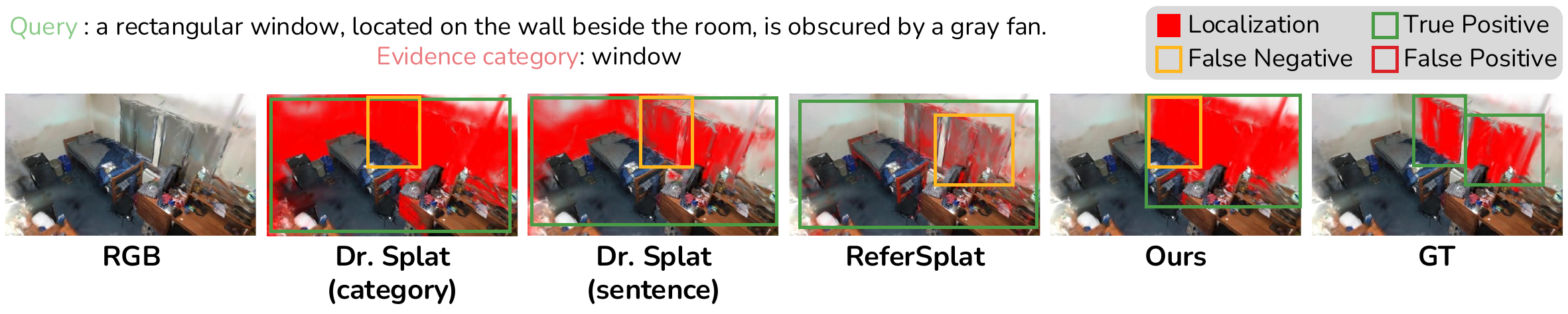}
    \caption{\textbf{Failure cases of Multi3DRefer (3D object localization) with multiple targets.}
    While accurate localization is challenging in itself, predicting the exact instance count remains difficult even when localization succeeds.
    Dr. Splat and ReferSplat do not explicitly produce a variable number of instance-separated predictions under our direct-3D evaluation protocol.
    In contrast, our method separates activations by instance, enabling instance-aware grounding, although some localization errors remain.
    }
    \label{fig:failure_case}
\end{figure}

\subsection{3D Referring Segmentation with Novel View Synthesis}
Due to the extensive scale of the ScanRefer~\cite{chen2020scanrefer} and Multi3DRefer~\cite{zhang2023multi3drefer} datasets, we restrict our evaluation of novel-view synthesis for 3D visual grounding to the 3D referring segmentation task. 
As shown in~\Tref{tab:referring_novel_view}, novel-view synthesis also improves performance in 3D visual grounding task.

\section{Limitation and Discussion}\label{supp:discussion}


A current limitation of our framework is its reliance on views near existing given camera viewpoints, leaving unconstrained optimal view exploration as future work.
Regarding failure cases, 3D object localization in multi-target settings presents a significant challenge. 
As illustrated in~\Fref{fig:failure_case}, accurately localizing multiple objects across varying quantities while determining their exact count remains a critical bottleneck, which serves as a key direction for future development.

\clearpage

\begin{figure}[t]
\centering
\begin{tcolorbox}[
    colback=gray!5,
    colframe=gray!75,
    boxrule=0.6pt,
    arc=2pt,
    left=5pt, right=5pt, top=5pt, bottom=5pt,
    width=\linewidth,
]
    \textbf{System Prompt:} \\
    You are an AI agent in a 3D indoor scene. \\[6pt]

    \textbf{Content Prompt:} \\
    You are an intelligent question answering agent. \\
    I will ask you questions about an indoor space and you must provide an answer. \\
    You will be shown a set of images that have been collected from a single location. \\
    Given a user query, you must output ``text'' to answer to the question asked by the user. \\

    Image 0 [img] \\
    Image 1 [img] \\
    ...\\[6pt]

    User Query: \{question\} \\[6pt]

\end{tcolorbox}
\caption{\textbf{Visual question-answering (visual QA) prompt in the VLM-as-Judge stage.}
The placeholders \{question\} and [img] are replaced with the user question and novel view images, respectively.
This prompt is for the initial answer prediction stage in~\Fref{fig:final_view_selection} of the main paper.
}
\label{fig:visual_qa}
\end{figure}
\begin{figure}[t]
\centering
\begin{tcolorbox}[
    colback=gray!5,
    colframe=gray!75,
    boxrule=0.6pt,
    arc=2pt,
    left=5pt, right=5pt, top=5pt, bottom=5pt,
    width=\linewidth,
]
    \textbf{System Prompt:} \\
    You are an AI agent in a 3D indoor scene. \\[6pt]

    \textbf{Content Prompt:} \\
    You are a reasoning agent that must select the most appropriate answer sentence to a given question. \\
    You are given a user query and a list of candidate sentences (each with an index). \\[6pt]
    
    1. Read the user query and all candidate sentences with their indices. \\
    2. For each sentence, judge how well it directly and specifically answers the query. \\
    3. Prefer sentences that provide a concrete, direct, and semantically correct answer to the user query. \\
    4. If exactly one sentence is clearly better matched than all others, you MUST return its index. \\
    5. Select the single sentence that best and most directly answers the user query. \\
    6. Always return the index of that sentence, even if several sentences are similar. \\

    User Query: \{question\} \\[6pt]

    Candidate Sentences: \\
    0: \{$\text{initial}\_\text{answer}\_\text{prediction}\_0$\} \\
    1: \{$\text{initial}\_\text{answer}\_\text{prediction}\_1$\} \\
    ... \\[6pt]

    Return: \\
    An integer index of the best answer sentence.

\end{tcolorbox}
\caption{\textbf{View selection prompt in the VLM-as-Judge stage.}
The placeholders \{question\} and \{$\text{initial}\_\text{answer}\_\text{prediction}$\} are replaced with the user question and predicted initial answer from the initial answer prediction stage, respectively.
This prompt is for the final view selection in~\Fref{fig:final_view_selection} of the main paper.
}
\label{fig:view_selection}
\end{figure}
\begin{figure}[t]
\centering
\begin{tcolorbox}[
    colback=gray!5,
    colframe=gray!75,
    boxrule=0.6pt,
    arc=2pt,
    left=5pt, right=5pt, top=5pt, bottom=5pt,
    width=\linewidth,
]
    \textbf{System Prompt:} \\
    You are an AI agent in a 3D indoor scene. \\[6pt]

    \textbf{Content Prompt:} \\
    You are an intelligent embodied question answering agent. \\
    You are a careful verification agent responsible for evaluating candidate answers to a user question and selecting the single most accurate and appropriate response. \\[6pt]

    Instructions: \\
    Pick ONE answer that explicitly and directly answers the question. \\
    Reject any that say they can't tell/are unsure (e.g., 'sorry', 'can't determine', 'not sure', 'maybe'). \\
    If one answer gives information about the object while another says it is not visible, prefer the one that gives the information. \\
    If tie, pick the lower index. \\

    User Question: \{question\} \\[6pt]

    Candidate Answers: \\
    0: \{$\text{candidate}\_\text{answer}\_0$\} \\
    1: \{$\text{candidate}\_\text{answer}\_1$\} \\[6pt]

    Return: \\
    An integer index of the best answer sentence. \\[6pt]

    Output Constraint: \\
    Output ONLY the integer index of the selected explanation. No extra text.

\end{tcolorbox}
\caption{\textbf{Verification prompt.}
The placeholder \{question\} is replaced with the user question.
\{$\text{candidate}\_\text{answer}\_0$\} denotes the answer predicted without the novel-view adjustment stage (i.e., only with initial view selection), \{$\text{candidate}\_\text{answer}\_1$\} denotes the answer predicted with the novel-view adjustment stage (i.e., with the final view selection).
The verification stage is crucial for determining whether the selected novel view reveals previously unobserved or occluded information that is essential for addressing the query.
This is for the verification stage in~\Fref{fig:final_view_selection} of the main paper.
}
\label{fig:verification}
\end{figure}
\begin{figure}[t]
\centering
\begin{tcolorbox}[
    colback=gray!5,
    colframe=gray!75,
    boxrule=0.6pt,
    arc=2pt,
    left=5pt, right=5pt, top=5pt, bottom=5pt,
    width=\linewidth,
]
    \textbf{System Prompt:} \\
    You are an AI agent in a 3D indoor scene. \\[6pt]

    \textbf{Content Prompt:} \\
    You are an intelligent question answering agent. \\
    I will ask you questions about an indoor space and you must provide an answer. \\
    You will be shown a set of images that have been collected from a single location. \\
    Given a user query, you must output ``text'' to answer to the question asked by the user. \\

    Image 0 [img] \\
    Image 1 [img] \\
    ...\\[6pt]

    User Query: \{question\} \\[6pt]

    Output Format: \\
    Please provide your answer in the following format: ``Image i \textbackslash n [Answer]'', where i is the index of the image you choose.
    For example, if you choose the first image, you can return ``Image 0 \textbackslash n The fruit bowl is on the kitchen counter.''

\end{tcolorbox}
\caption{\textbf{Prompt of embodied question-answering and 3D visual grounding (single target).}
The placeholders \{question\} and [img] are replaced with the user question and novel view images, respectively.
We append an output-format constraint that requires the VLM to specify which image is responsible for its answer.
The selected index identifies the corresponding camera viewpoint, which is then used for 3D object localization on ScanRefer and 3D referring segmentation, where each query refers to a single target instance.
This prompt predicts the definitive answer used for calculating LLM-Match scores.
}
\label{fig:eqa}
\end{figure}
\begin{figure}[t]
\centering
\begin{tcolorbox}[
    colback=gray!5,
    colframe=gray!75,
    boxrule=0.6pt,
    arc=2pt,
    left=5pt, right=5pt, top=5pt, bottom=5pt,
    width=\linewidth,
]
    \textbf{System Prompt:} \\
    You are an AI agent in a 3D indoor scene. \\[6pt]

    \textbf{Content Prompt:} \\
    You are an intelligent visual grounding agent. \\
    I will describe an object or objects in an indoor space, and you must identify which images contain matching target object instances. \\
    You will be shown a set of images that have been collected from a single location. \\
    Given the user query, decide independently for each image whether it contains a matching target object. \\
    There may be zero, one, or multiple matching images. \\
    Use 0-based image indices. The first image is Image 0, the second image is Image 1, and so on. \\

    Image 0 [img] \\
    Image 1 [img] \\
    ...\\[6pt]

    User Query: \{question\} \\[6pt]

    Output Format: \\
    For every image, return exactly one line in this format: \\
    `Image i: Yes' or `Image i: No'. \\
    Use `Yes' only if that image contains a matching target object instance. Use `No' if the target is not visible or does not match. \\
    Use 0-based image indices exactly as provided above. Do not use 1-based numbering and do not start from Image 1. \\
    Do not return JSON, markdown, explanations, or any text outside the image lines. \\
    Example:\textbackslash{}nImage 0: No\textbackslash{}nImage 1: No\textbackslash{}nImage 2: Yes\textbackslash{}n

\end{tcolorbox}
\caption{\textbf{Prompt of 3D visual grounding (multiple targets).}
The placeholders \{question\} and [img] are replaced with the user question and novel view images, respectively.
We append an output-format constraint that requires the VLM to specify which image is responsible for its answer.
The selected index is then used to determine the corresponding camera, from which we perform 3D visual grounding.
}
\label{fig:grounding}
\end{figure}

\end{document}